\documentclass[conference]{IEEEtran}
\IEEEoverridecommandlockouts
\usepackage{cite}
\usepackage{amsmath,amssymb,amsfonts}
\usepackage{algorithmic}
\usepackage{graphicx}
\usepackage{textcomp}
\usepackage{xcolor}
\usepackage{booktabs}

\usepackage{listings}
\usepackage{pgfplots,pgfplotstable}
\usetikzlibrary{patterns.meta}
\usetikzlibrary{positioning,fit,arrows.meta,backgrounds}
\usetikzlibrary{backgrounds}
\usepackage{pgf-pie} 

\pgfplotsset{compat=newest}

\definecolor{dorange}{RGB}{204,102,0}
\definecolor{cyan}{RGB}{0,153,255}

\definecolor{blue0}{RGB}{0,0,150}
\definecolor{blue1}{RGB}{30,60,255}
\definecolor{blue2}{RGB}{60,120,255}
\definecolor{blue3}{RGB}{0,184,230}
\definecolor{blue4}{RGB}{194,194,255}
\definecolor{blue5}{RGB}{38,77,115}

\tikzset{
    module/.style={%
        draw, rounded corners,
        font=\sffamily
        },
    module/.default=2cm,
    >=LaTeX
}

\makeatletter
\tikzset{
    database/.style={
        path picture={
            \draw (0, 1.5*\database@segmentheight) circle [x radius=\database@radius,y radius=\database@aspectratio*\database@radius];
            \draw (-\database@radius, 0.5*\database@segmentheight) arc [start angle=180,end angle=360,x radius=\database@radius, y radius=\database@aspectratio*\database@radius];
            \draw (-\database@radius,-0.5*\database@segmentheight) arc [start angle=180,end angle=360,x radius=\database@radius, y radius=\database@aspectratio*\database@radius];
            \draw (-\database@radius,1.5*\database@segmentheight) -- ++(0,-3*\database@segmentheight) arc [start angle=180,end angle=360,x radius=\database@radius, y radius=\database@aspectratio*\database@radius] -- ++(0,3*\database@segmentheight);
        },
        minimum width=2*\database@radius + \pgflinewidth,
        minimum height=3*\database@segmentheight + 2*\database@aspectratio*\database@radius + \pgflinewidth,
    },
    database segment height/.store in=\database@segmentheight,
    database radius/.store in=\database@radius,
    database aspect ratio/.store in=\database@aspectratio,
    database segment height=0.1cm,
    database radius=0.25cm,
    database aspect ratio=0.35,
}
\makeatother

\definecolor{rosso}{RGB}{220,57,18}
\definecolor{giallo}{RGB}{255,153,0}
\definecolor{blu}{RGB}{102,140,217}
\definecolor{verde}{RGB}{16,150,24}
\definecolor{viola}{RGB}{153,0,153}

\makeatletter

\tikzstyle{chart}=[
    legend label/.style={font={\scriptsize},anchor=west,align=left},
    legend box/.style={rectangle, draw, minimum size=5pt},
    axis/.style={black,semithick,->},
    axis label/.style={anchor=east,font={\tiny}},
]

\tikzstyle{bar chart}=[
    chart,
    bar width/.code={
        \pgfmathparse{##1/2}
        \global\let\bar@w\pgfmathresult
    },
    bar/.style={very thick, draw=white},
    bar label/.style={font={\bf\small},anchor=north},
    bar value/.style={font={\footnotesize}},
    bar width=.75,
]

\tikzstyle{pie chart}=[
    chart,
    slice/.style={line cap=round, line join=round, very thick,draw=white},
    pie title/.style={font={\bf}},
    slice type/.style 2 args={
        ##1/.style={fill=##2},
        values of ##1/.style={}
    },
    slice type pattern/.style 2 args={
        ##1/.style={pattern=##2},
        values of ##1/.style={}
    }
]

\pgfdeclarelayer{background}
\pgfdeclarelayer{foreground}
\pgfsetlayers{background,main,foreground}
\usetikzlibrary{patterns}

\newcommand{\legend}[2][]{
    \begin{scope}[#1]
    \path
        \foreach \n/\s in {#2}
            {
                  ++(0,-10pt) node[\s,legend box] {} +(5pt,0) node[legend label] {\n}
            }
    ;
    \end{scope}
}

\usepackage{scalerel}
\usepackage{tikz}
\usetikzlibrary{svg.path}

\definecolor{orcidlogocol}{HTML}{A6CE39}
\tikzset{
  orcidlogo/.pic={
    \fill[orcidlogocol] svg{M256,128c0,70.7-57.3,128-128,128C57.3,256,0,198.7,0,128C0,57.3,57.3,0,128,0C198.7,0,256,57.3,256,128z};
    \fill[white] svg{M86.3,186.2H70.9V79.1h15.4v48.4V186.2z}
                 svg{M108.9,79.1h41.6c39.6,0,57,28.3,57,53.6c0,27.5-21.5,53.6-56.8,53.6h-41.8V79.1z M124.3,172.4h24.5c34.9,0,42.9-26.5,42.9-39.7c0-21.5-13.7-39.7-43.7-39.7h-23.7V172.4z}
                 svg{M88.7,56.8c0,5.5-4.5,10.1-10.1,10.1c-5.6,0-10.1-4.6-10.1-10.1c0-5.6,4.5-10.1,10.1-10.1C84.2,46.7,88.7,51.3,88.7,56.8z};
  }
}

\newcommand\orcidicon[1]{\href{https://orcid.org/#1}{\mbox{\scalerel*{
\begin{tikzpicture}[yscale=-1,transform shape]
\pic{orcidlogo};
\end{tikzpicture}
}{|}}}}

\usepackage{hyperref} 

\tikzdeclarepattern{
  name=mylines,
  parameters={
      \pgfkeysvalueof{/pgf/pattern keys/size},
      \pgfkeysvalueof{/pgf/pattern keys/angle},
      \pgfkeysvalueof{/pgf/pattern keys/line width},
  },
  bounding box={
    (0,-0.5*\pgfkeysvalueof{/pgf/pattern keys/line width}) and
    (\pgfkeysvalueof{/pgf/pattern keys/size},
0.5*\pgfkeysvalueof{/pgf/pattern keys/line width})},
  tile size={(\pgfkeysvalueof{/pgf/pattern keys/size},
\pgfkeysvalueof{/pgf/pattern keys/size})},
  tile transformation={rotate=\pgfkeysvalueof{/pgf/pattern keys/angle}},
  defaults={
    size/.initial=5pt,
    angle/.initial=45,
    line width/.initial=.4pt,
  },
  code={
      \draw [line width=\pgfkeysvalueof{/pgf/pattern keys/line width}]
        (0,0) -- (\pgfkeysvalueof{/pgf/pattern keys/size},0);
  },
}

\def\BibTeX{{\rm B\kern-.05em{\sc i\kern-.025em b}\kern-.08em
    T\kern-.1667em\lower.7ex\hbox{E}\kern-.125emX}}
\begin{document}

\title{gSuite: A Flexible and Framework Independent Benchmark Suite for Graph Neural Network Inference on GPUs}

\author{
\IEEEauthorblockN{Taha Tekdoğan\IEEEauthorrefmark{1}\textsuperscript{,}\IEEEauthorrefmark{2} \orcidicon{0000-0003-2091-3192}, 
Serkan Göktaş\IEEEauthorrefmark{1} \orcidicon{0000-0002-4701-3090}, and 
Ayse Yilmazer-Metin\IEEEauthorrefmark{1} \orcidicon{0000-0003-4502-7365}}
\\
\IEEEauthorblockA{\IEEEauthorrefmark{1}\textit{Department of Computer Engineering}, 
\textit{Istanbul Technical University},
Istanbul, Turkey \\}
\IEEEauthorblockA{\IEEEauthorrefmark{2}\textit{Radar and Electronic Warfare Intelligence Systems}, 
\textit{ASELSAN Inc.},
Ankara, Turkey \\}
\\
Email: \IEEEauthorrefmark{1}\{tekdogan20,
goktas16,
yilmazerayse\}@itu.edu.tr
}

\maketitle

\begin{abstract}

As the interest to Graph Neural Networks (GNNs) is growing, the importance of benchmarking and performance characterization studies of GNNs is increasing.
So far, we have seen many studies that investigate and present the performance and computational efficiency of GNNs.
However, the work done so far has been carried out using a few high-level GNN frameworks.
Although these frameworks provide ease of use, they contain too many dependencies to other existing libraries.
The layers of implementation details and the dependencies complicate the performance analysis of GNN models that are built on top of these frameworks, especially while using architectural simulators.
Furthermore, different approaches on GNN computation are generally overlooked in prior characterization studies, and merely one of the common computational models is evaluated.
Based on these shortcomings and needs that we observed, we developed a benchmark suite that is framework independent, supporting versatile computational models, easily configurable and can be used with architectural simulators without additional effort.

Our benchmark suite, which we call gSuite, makes use of only hardware vendor's libraries and therefore it is independent of any other frameworks.
gSuite enables performing detailed performance characterization studies on GNN Inference using both contemporary GPU profilers and architectural GPU simulators.
To illustrate the benefits of our new benchmark suite, we perform a detailed characterization study with a set of well-known GNN models with various datasets; running gSuite both on a real GPU card and a timing-detailed GPU simulator.
We also implicate the effect of computational models on performance.
We use several evaluation metrics to rigorously measure the performance of GNN computation.
We make gSuite available to research community and provide all the configuration settings which we used for our evaluation so that all the experiments mentioned in the paper are reproducible.
\end{abstract}


\section{Introduction}
Graph structured data are highly preferred in many real-world applications due to their ability of expressing the topology of irregular domains.
For instance, graphs are used for representing molecules in chemistry\cite{Gilmer2017}, relationships among people in social sciences\cite{Nettleton2013}, and connections between brain areas in computational neuroscience\cite{Li2020}.
Real-world graph datasets have been scaled to enormous amount of sizes in terms of number of nodes, edges, and their feature lengths.
Processing these huge-sized data requires an intensive computation.
Utilizing Graphics Processing Units (GPUs) is the de facto method in order to meet computation requirements of the graph operations.

Successful application of deep learning techniques in many areas has triggered the idea of applying deep neural network(DNN)-based techniques on the graph structured data.
Graph Neural Networks (GNNs) are deep learning based methods that are capable of working on non-euclidean data.
GNNs provide a way of performing node level, edge level, and graph level prediction for graph structured data.

There have been various approaches to carry out GNN operations such as message passing (MP)\cite{Gilmer2017,Gilmer2020} and sparse matrix multiplication (SpMM)\cite{Ashari2014}.
Therefore, GNN computation can be applied in various ways.
The increasing number of GNN research motivated developers to extend commonly used deep learning frameworks to support GNN operations using MP or SpMM computational models.

The most popular GNN frameworks are built on top of the commonly used Python-based deep learning frameworks (e.g., PyTorch Geometric (PyG)\cite{MatthiasFey2019} is built on PyTorch; Deep Graph Library (DGL)\cite{Wang2019} gives end user a choice to alternate among PyTorch, Tensorflow or MXNet).
Even though these frameworks provide ease of development, they bring dependency to implementations within the base-lined framework and the underlying development libraries.

The increasing number of studies in GNN area has led the benchmarking studies to evaluate the performance of GNN computations.
In Table~\ref{tab:studies}, we summarize the main frameworks and benchmarking studies on GNNs.
As the table shows, all of the existing frameworks and benchmarking studies utilize at least one of the existing DNN/GNN frameworks and their libraries.
The dependencies to the existing frameworks rather complicates the performance analysis and characterization of GNN computations.
Most of the computer architecture studies favor utilizing detailed architectural simulators.
However, the dependency chain of the existing frameworks makes performance analysis of the GNN applications inaccessible, especially while using architectural simulators. 

While the GNN frameworks are intended to be extendable, benchmarks and characterization studies have been only on a limited number of well known GNN models and datasets\cite{gcn-char,gnn-char,gnnmark}.
Additionally, most of these frameworks and benchmarks are based on a specific computational model.

All of these limitations of the existing studies and efforts motivated us to develop a configurable and framework independent benchmark suite for GNN Inference.
In this paper, we introduce our GNN benchmark suite which we call \emph{gSuite}.
gSuite is highly flexible, allows either using an existing framework (such as PyG or DGL) or using our GNN implementations that only make use of the hardware vendor's libraries.
The parameters of the desired GNN pipeline, such as the GNN model, the dataset, the number of GNN layers, etc., can be easily configured by passing a few parameters to the program.
We built gSuite as the collection of utilities (data import, transform, etc.) and core kernels which are the most primitive operations of GNNs.
Therefore, it is extendable to create new GNN models and study their performance on GPUs.
gSuite is designed for efficiently studying the performance of GNNs with either hardware profilers or cycle accurate simulators.
It does not require additional effort to utilize an architectural simulator, which makes GNN-related operations quite accessible in terms of performance characterization.
As a proof of concept, in this work, we characterize the most popular GNN models on varying datasets.
We interpret the experimental results in terms of the effect of input workload, GNN model and computational model.

In summary, in this study, we make the following contributions:
\begin{itemize}
    \item We provide a flexible and user-friendly benchmark suite for GNN inference, hence a desired GNN pipeline can be easily built by passing only a few parameters.
    \item We eliminate the dependency of GNN pipelines to other frameworks (such as PyG and DGL) and machine learning utilities (such as PyTorch and Tensorflow).
    \item We characterize the computation of the most representative GNN models at inference level by comparing two predominant computational models, i.e., message passing (MP) and sparse matrix multiplication (SpMM).
    \item We demonstrate the accessibility of GNN performance on gSuite using both a hardware profiler and an architectural simulator.
    \item We present and summarize our performance results and make architectural suggestions based on our findings.
\end{itemize}

Remaining of this paper is structured as follows:
Section \ref{background} introduces fundamentals of GNNs with their notations, formulas, common data formats, prevalent datasets, and popular GNN frameworks.
In Section \ref{sota}, we discuss the state-of-the-art of benchmarking methodologies and characterization studies for GNNs.
Section \ref{ourMethodology} explains our architectural model by declaring core kernels as the most primitive GNN operations.
And finally in Section \ref{evaluation}, we deliver our benchmarking methodology for evaluating the performance of the GNN computation.
Then we deliver our results and discuss them in detail.

\section{A Brief Background on GNNs}\label{background}
%
GNNs were first introduced by Scarselli et al.\cite{Scarselli2009}, and many new GNN models were proposed since then\cite{Chen2018, Velickovic2018,Li2018, Zhang_Cui_Neumann_Chen_2018}.
A large set of domains leverage the capability of GNNs\cite{Gilmer2017,Nettleton2013,Li2020}.
Adopting GNNS to wide range of domains gives rise to many new GNN models with various characteristics.
Below, we provide a brief background on GNNs by introducing common notation, the computational approaches and popular frameworks to implement the required GNN operations, widely used graph datasets and graph formats utilized to express them.

Graphs are widely used fundamental data structures that are very successful at expressing real-world data that includes relationships between its entities.
A graph $G = (V,E)$ is defined by a set of nodes $V$, and a set of edges $E$.
Two nodes are neighbours if they are directly connected to each other with an edge.
The set of neighbour nodes of a node $v$ is represented as $N(v)$.
Nodes may carry a list of features that are represented with a latent vector which holds information, known as \emph{node embedding} in GNN literature.
We represent the node embedding of a node $v$ as $h_v$.

GNN pipelines generally consist of multiple GNN layers $L$.
We denote a specific node embedding for layer $k$ as $h^{(k)}_v$, where $k~\epsilon~[1,L]$ stands for current GNN layer.
In some cases, edges may carry information which is called an \emph{edge embedding}, and it is represented as $g_e$.
Most often, the task of a GNN model is predicting or generating the node or edge embeddings. 

Furthermore, node and the edge embeddings can be represented with matrices, instead of latent vectors.
Feature information of the vertices in a graph can be represented with a feature matrix $X$ in shape $[|V|,f]$, where $f$ stands for the feature size.
The connectivity information between nodes in a graph can also be represented with an adjacency matrix $A$ in shape $[|V|,|V|]$.

We will be using these notations for explaining the mathematical expressions of a core set of GNN models that are widely used and also implemented in our benchmark suite.
Table ~\ref{tab:notations} summarizes this notation that we use during our study.

\begin{table}[]
    \centering
    \caption{Notations for graph neural networks}
    \begin{tabular}{cl}
        \toprule
        \textbf{Notation} & \textbf{Description} \\
        \midrule
        $G(V,E)$    & Graph \\
        $V$         & Set of nodes of the graph \\
        $|V|$       & Number of nodes in the graph \\
        $E$         & Set of edges of the graph \\
        $v$         & A single node where $v~\epsilon~V$ \\
        $e$         & A single edge where $e~\epsilon~E$ \\
        $k$         & Current GNN layer where $k~\epsilon~[1,L]$ \\
        $h^{(k)}_v$ & Feature representation of node $v$ at layer $k$ \\
        $g^{(k)}_e$ & Feature representation of edge $e$ at layer $k$ \\
        $N(v)$      & Neighbourhood nodes of the node $v$ \\
        $A$         & Adjacency matrix of the graph \\
        $X^{(k)}$   & Feature matrix of the graph at layer $k$ \\
        \bottomrule
    \end{tabular}
    \label{tab:notations}
\end{table}

\subsection{Computation of GNNs}
GNN computation can be evaluated under two major GNN phases: inference and training.
Inference phase refers to updating each node embedding in a graph analogous to corresponding GNN schema and pre-trained model coefficients.
Training phase refers to optimizing the coefficients of the model.
Here in this work, we mainly focus on the inference stage of GNNs.
Therefore when we invoke the GNN computation, we imply the computation of GNN inference during the study.

A typical GNN includes two types of operations: \emph{aggregation} (or \emph{message} in some cases \cite{MatthiasFey2019, Gilmer2017, Wang2019}) and \emph{combination} (or \emph{update} in some cases \cite{Peter2018, MatthiasFey2019, Gilmer2017,Liao2018, Wang2019}).
\emph{Aggregation} refers to capturing information from a node's neighbour nodes and accumulating them into its feature representation.
It is done by a predefined aggregator function such as $sum$, $mean$, and $max$.
\emph{Combination} refers to updating a node's representation by using the output of aggregation phase, which is mostly a multilayer perceptron (MLP)\cite{Marius2009}.
Aggregation and combination operations are applied analogous to definition of the corresponding GNN model.
Application of these operations forms the implementation of the mathematical definitions of GNN models.

Aggregation and combination operations can be applied to graph datasets based on two classes of computational models: \emph{Message Passing} (MP)\cite{Gilmer2017,Gilmer2020} and \emph{Sparse Matrix Multiplication} (SpMM)\cite{Wang2019}.
\emph{MP} model is based on a computation pattern where connected nodes scatter their attributes through neighbourhood nodes (aggregation) and each node updates its node-embedding by using such neighborhood nodes' features (combination).
On the other hand, \emph{SpMM} model refers to applying aggregation and combination schemes by reducing them into a sequence of matrix multiplication operations.

\subsection{GNN Frameworks}
There are a number of frameworks to provide an infrastructure to build and run GNNs pipelines such as PyTorch Geometric\cite{MatthiasFey2019}, Deep Graph Library\cite{Wang2019}, Graph Nets\cite{GraphNets}, and Spektral\cite{Spektral}.
PyTorch Geometric (PyG) and Deep Graph Library (DGL) are the most popular ones among all these frameworks.
PyG is built on top of PyTorch library, and all the implemented GNN models are inherited from a base class called \emph{MessagePassing}.
On the other hand, DGL implements GNN models based on SpMM computational model.
It gives user a choice to alternate between three frameworks (PyTorch, Tensorflow and MXNet).


We examined widely used GNN frameworks and their model implementations.
Then we imitated the MP kernels from PyG and SpMM kernels from DGL to implement core kernels of GNNs.
We also purified these kernels from dependencies to other libraries (such as PyTorch).
Table~\ref{tab:kernels} provides the list of identified core MP and SpMM kernels.

\begin{table}
  \caption{Core MP and SpMM kernels.}
  \label{tab:kernels}
  \begin{tabular}{cccl}
    \toprule
    \textbf{Kernel}  & \textbf{Computational}  & \textbf{Short} & \textbf{Description}\\
    \textbf{Name}     & \textbf{Model}         & \textbf{Form} \\
    \midrule
    \textbf{indexSelect}        & MP     & is & Indexes the input along\\
                                &        &    & specified dimension by \\
                                &        &    & using index entries.\\
    \midrule
    \textbf{scatter}            & MP     & sc & Reduces given input \\
                                &        &    & based-on index vector \\
                                &        &    & using entries. \\
    \midrule
    \textbf{sgemm}              & SpMM   & sg & Generalized matrix \\
                                & /GEMM  &    & multiplication of \\
                                &        &    & two given matrices. \\
    \midrule
    \textbf{SpGEMM}             & SpMM   & sp & Matrix multiplication\\
                                & /GEMM  &    & of two sparse matrices. \\
    
  \bottomrule
\end{tabular}
\end{table}

MP models generally consist of neighbour node calculation (\emph{indexSelect}), scattering the node embedding through these connections (\emph{scatter}), and updating self node embedding with a linear function (\emph{sgemm}).
On the other hand, SpMM models consist of a consecutive execution of matrix multiplication operations (\emph{SpGEMM} and \emph{sgemm}).
These core kernels are organized to comply corresponding GNN model's computation formula.


\subsection{GNN Models}
While it is very easy to extend our benchmark suite to include any type of GNN model, we have chosen three widely-used GNN models to implement and base our discussions in this paper.
We demonstrate the implementation and detailed performance characterization of these three GNN models using gSuite.
These three GNN models are \emph{Graph Convolutional Network (GCN)} \cite{semi-supervised-gcn}, \emph{Graph Isomorphism Network (GIN)} \cite{Xu2019}, and \emph{GraphSAGE (SAG)} \cite{Hamilton2017}.
Using the notation that we presented above, we continue providing implementation details of these three widely-used GNN models.
Then we explain the implementation of the core kernels of GNNs by mapping the formulas with computational models.

\subsubsection{Graph Convolutional Networks}\label{gcn}
\emph{Graph Convolutional Network (GCN)} is a semi-supervised classification method that is an efficient variant of convolutional neural networks designed to operate on graphs\cite{semi-supervised-gcn}. It is motivated by the idea of using layer-wise propagation on graph structured data. GCN is capable of encoding both node features and graph structure with their proposed graph modeling approach. Therefore, it is quite popular in a wide range of implementations from knowledge embedding\cite{Yao2022} to face clustering\cite{Sun2022}.

We can express the GCN computation using both \emph{MP} and \emph{SpMM} computational model.
The message passing formula for updating each node embedding of a graph in GCN is given by \eqref{eq_gcn_mp}.

\begin{equation}
\label{eq_gcn_mp}
    {h_v^{(k+1)}} = {\Theta}\Bigg( \sum_{u\epsilon N(v)\cup\{v\}} \frac{1}{\sqrt{d_u d_v}} h_u \Bigg)
\end{equation}

In \eqref{eq_gcn_mp}, ${h_v}^{(k+1)}$ is the feature representation of the updated node $v$ in ${(k+1)}^{th}$ layer.
$h_u$ is a neighbour node embedding from the set $u\epsilon N(v)\cup\{v\}$.
$d_v$ represents the node degree of node $v$, i.e. the number of edges connected to node $v$.
$\Theta$ is a linear activation function.

The formulation of GCN using SpMM model is given in \eqref{eq_gcn_spmm}.

\begin{equation}
\label{eq_gcn_spmm}
    {X^{k+1}} = {\hat{D}}^{-1/2}{\hat{A}}{\hat{D}}^{-1/2}X^{k}\Theta
\end{equation}

In \eqref{eq_gcn_spmm}, $X^{(k+1)}$ represents the feature matrix of a graph at layer $(k+1)$.
$\hat{A}$ is the adjacency matrix with self-loops inserted, i.e.:
\begin{displaymath}
  \hat{A} = A + I.
\end{displaymath}
$\hat{D}$ is $\hat{A}$'s diagonal matrix, i.e.:
\begin{displaymath}
  \hat{D}_{ii} = \sum_{j=0}\hat{A}_{ij}.
\end{displaymath}
And finally, $\Theta$ is an activation function such as a Rectified Linear Unit (ReLU)\cite{Glorot2011} or a Sigmoid function\cite{Narayan1997}.

\subsubsection{Graph Isomorphism Networks}
\normalsize \emph{Graph Isomorphism Networks (GINs)} combine the discriminative power of Weisfeiler-Lehman (WL) graph isomorphism test \cite{Xu2019, Weisfeiler1968} with GNN's recursive neighbourhood aggregation by making aggregation phase highly expressive and modeling injective functions. GINs are mostly used for classification and discrimination tasks on graphs\cite{Bandyopadhyay2021}. Following formulas explain how node embeddings are updated using \emph{MP} and \emph{SpMM} computational approaches for implementing GINs. \eqref{gin-mp} shows the MP formula of a single GIN layer, and \eqref{gin-mm} shows the matrix multiplication version of a GIN layer computation.

\begin{equation}
    \centering\large {h_v^{(k)}} = {\Theta^{(k)}}\Bigg({(1+\epsilon^{(k)})} \ast {h_v^{(k-1)}} + \sum_{u\epsilon N(v)} {h_u^{(k-1)}}\Bigg)\normalsize
\label{gin-mp}
\end{equation}

In \eqref{gin-mp}, $h_v^{(k)}$ represents the feature vector of the node $v$ at layer $k$.
$h_u^{(k-1)}$ is the feature representation of a neighbourhood node $u$ at layer $(k-1)$.
$\varepsilon$ is a constant, and $\Theta^{(k)}$ is an activation function at layer $k$.

\begin{equation}
    X^{k+1} = \Theta^{(k)}\Bigg({\Big(A^{k} + (1 + \epsilon)\cdot I\Big) \cdot X^{k}}\Bigg)
\label{gin-mm}
\end{equation}

In \eqref{gin-mm}, $X^{k+1}$ represents the feature matrix of a graph.
$A$ is an adjacency matrix, $I$ is an identity matrix, $\epsilon$ is a constant, and $\Theta^{(k)}$ is an activation function at layer $k$.

\subsubsection{GraphSAGE}
\emph{GraphSAGE (SAG)} is a general inductive model which generates previously unseen nodes in a graph by leveraging the current node information \cite{Hamilton2017}.
SAG uses aggregating functions to aggregate feature information from node's local neighborhood, instead of training a distinct embedding vector for each node. Even though we could not find an available SpMM version of SAG, we implemented it using only the MP computational model due to its popularity with unsupervised learning on graph structured data\cite{Chen2022, Zhang2022}.

Equation~\eqref{sag-mp} shows the formula of MP-oriented SAG model.

\begin{equation}
\label{sag-mp}
    \centering\large {h_v^{(k)}} = W_1h_v^{(k-1)} + W_2\ast mean_{j\epsilon N(v)\cup\{v\}}h_u\normalsize
\end{equation}

In \eqref{sag-mp}, $h_v^{(k)}$ shows the feature representation of a node $v$ at layer $k$.
$W_1$ and $W_2$ are scalar weights for self nodes and neighbour nodes, respectively.


\subsection{Datasets and Widely Used Graph Formats}

In prior GNN studies, we often see datasets that consist of two parts: connectivity information to represent edges in graphs, and content information to embody node embeddings.
Frameworks construct graphs in terms of their utilized graph formats by inferring information from these datasets.
The most popular graph datasets in GNN studies are Cora, Citeseer\cite{cora}, Pubmed\cite{pubmed}, Reddit\cite{reddit} and LiveJournal\cite{LJdataset1,LJdataset2}.

Graph datasets are generally transformed to one of the following formats to be processed by graph libraries: dense matrix, sparse matrix, coordinate format (COO) and compressed sparse row (CSR).

Dense and sparse matrices are often used as input to graph operations based on matrix multiplication, such as SpMM.
On the other hand, COO and CSR formats are compressed formats of the graphs that represent attributes and topology of the graph in low-dimensional vector.
These types of graph data formats are commonly used in MP-based frameworks, such as PyG.
We include all of these formats in our work, and provide utilities to transform a dataset from one format to another.



\section{Limitations of Existing GNN Frameworks and Benchmarking Efforts}\label{sota}




\begin{table*}

  \begin{center}

  \caption{Summary of the prior GNN frameworks, benchmarks, characterization studies and gSuite with their properties.}
  
  \begin{tabular}{lcccccc}
  
    \toprule
    \textbf{Study Name}     & \textbf{GNN Models}  & \textbf{Frameworks}    & \textbf{Datasets}  & \textbf{Extendibility} & \textbf{GNN Scope} \\
    \midrule
    
    \textbf{Pytorch Geometric}\cite{MatthiasFey2019}  & GCN, SAG, GIN,           & Pytorch & Cora, CiteSeer,  & Yes & Both  \\
                                                      & RGCN, ...                &         & Pubmed, MUTAG,   &  \\
                                                      &                          &         & PROTEINS, ... \\
    \\
                                                      
    \textbf{Deep Graph Library}\cite{Wang2019}        & GCN, GAT, SAG,   & Pytorch, MXNet, & REDDIT, ARXIV, & Yes & Both \\
                                                      & GIN, SGC, ...    & Tensorflow      & PROTEINS, ... \\

    \\
    
    \textbf{GCN-GPU}                                  & GCN, GIN, SAG     & PyG & Cora, CiteSeer,       &   No & Inference \\
    \textbf{Characterization}\cite{gcn-char}          &                   &     &  Pubmed, Reddit,      \\
                                                      &                   &     &  LiveJournal          \\
                                                      &                   &     &                       \\
    \\                                                  
                                                      
    \textbf{GNN-GPU}                                  & GCN, GAT, GGNN,   & PyG, DGL   & Cora, CiteSeer,  & No & Inference \\
    \textbf{Characterization}\cite{gnn-char}          &                   &            & Pubmed, AIFB,    \\
                                                      &                   &            & MUTAG, BGS       \\
                                                      &                   &            &                  \\
    \\
    
    \textbf{GNNMark}\cite{gnnmark}                     & PinSAGE, STGCN,    & PyG   & Cora, CiteSeer,   & No & Training \\
                                                       & DGCN, GW, KGNN,    &       & Pubmed, NWP,      \\
                                                       & ARGA, TLSTM        &       &  MVL, LA, PEMS    \\
                                                       &                    &       &                   \\
    \\

    \textbf{HyGCN}\cite{hygcn}                        & GCN, SAG, GIN    & PyG   & IMDB, Cora,      & No & Inference \\
                                                      &                  &       & Citeseer, Colab, \\
                                                      &                  &       &  Pubmed, Reddit  \\
                                                      &                  &       &                  \\
    \\
    
    \textbf{GRIP}\cite{grip}                            & GCN, GIN,     & GReTa & Pokec, YouTube        & No & Inference \\
                                                        & G-GCN, SAG    &       & LiveJournal, Reddit   \\
    
    \\
                                                        
    \textbf{gSuite}                                 &  \textbf{GCN, GIN, SAG}   & \textbf{None}  & \textbf{Cora, Citeseer, Pubmed}   & \textbf{Yes}   & \textbf{Inference} \\
                                                    &                           &                & \textbf{Reddit, LiveJournal} \\

    
    
    
    
  \bottomrule
  
\end{tabular}
\label{tab:studies}

\end{center}
\end{table*}

As the GNNs are finding application in many areas; several new frameworks, performance analysis and characterization studies have emerged.
We review the prior GNN frameworks, benchmarks and characterization studies in chronological order and evaluate them in terms of configurability, framework dependency, model and dataset versatility.
A comparison table Table~\ref{tab:studies} is provided to show existing studies' capabilities in terms of measuring GNN performance and ease of use.

As the Table~\ref{tab:studies} summarizes, prior works lack one or more of the attributes that is desired for studying the performance characteristics of GNN applications.
All of these studies utilize an existing DNN/GNN framework and have layers of dependency chain.
Such dependency may decrease the accessibility of GNN performance, especially while utilizing an architectural GPU simulator.

Furthermore, most of the frameworks, benchmarks and characterization studies assume that there merely exists a single computational approach.
For instance, Pytorch Geometric (PyG)\cite{MatthiasFey2019} follows a MP schema as a base class to whole GNN models.
On the other hand, Deep Graph Library (DGL)\cite{Wang2019} considers GNN computation as an SpMM problem.
Benchmarks and characterizations studies generally utilize one of these frameworks to build GNN pipelines.
As a result, such assumption on computational model may limit or lead to wrong conclusions when studying performance characteristics of the workloads.

Moreover, except the GNN development frameworks, the rest of the studies are not extendable.
One cannot create a new model or add a specific dataset.

With this study, we identify the need for a benchmark suite that does not limit the users to perform a thorough architectural performance analysis study. 

\section{gSuite and Our Design Approach}\label{ourMethodology}
While developing our GNN benchmark suite, we considered three key features: (1) Flexibility, (2) Extendability, and (3) Independence.
These features are explained below to point out the cornerstones of our benchmark suite's design approach.

\begin{itemize}

\item gSuite is a collection of utility functions (e.g. functions to allow input/output, setting configuration, etc.) and core kernels of MP and SpMM computational models.
It is flexible to allow building GNN pipelines by selecting the desired dataset, GNN model, number of layers, computational model, and framework (using either gSuite's core kernels or other framework's implementations).

\item gSuite allows researchers and engineers to extend the suite in any direction.
By utilizing MP and SpMM core kernels, a new GNN model can be built in a plug-and-play manner.

\item gSuite's core kernels do not have any dependency on any GNN/DNN frameworks.
However, we still give a choice to the end user for alternating between a GNN framework (PyG or DGL) and our GNN implementations.

\end{itemize}

\subsection{Software Architecture}
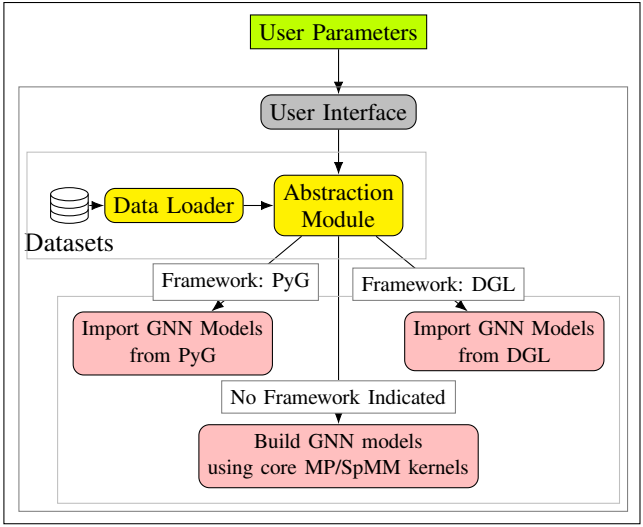
\begin{figure}[h]
  \centering

\begin{tikzpicture}[
 show background rectangle]

 \node[rectangle, draw, pattern=horizontal lines, fill=lime, font=\small]
 (I1) {User Parameters};
 \node[module, below=6mm of I1, pattern=crosshatch dots, fill=lightgray, font=\small]
 (I2) {User Interface};
 \node[module, below=6mm of I2, pattern=north west lines, fill=yellow, font=\small, align=center]
 (I3) {Abstraction\\Module};
 \node[module, left=4mm of I3, pattern=north west lines, fill=yellow, font=\small]
 (I10) {Data Loader};
 \node[module=1cm, below left=10mm and 0.1mm of I3, pattern=vertical lines, fill=pink, font=\small, align=center]
 (I5) {\footnotesize Import GNN Models\\ \footnotesize from PyG};
 \node[module=1cm, below=25mm of I3, pattern=vertical lines, fill=pink, font=\small, align=center]
 (I4) {\footnotesize Build GNN models\\ \footnotesize using core MP/SpMM kernels};
 \node[module=1cm, below right=10mm and 0.1mm of I3, pattern=vertical lines, fill=pink, font=\small, align=center]
 (I6) {\footnotesize Import GNN Models\\ \footnotesize from DGL};
 \node[database, label=below:Datasets,left= 2mm of I10] (D1) {};

 \foreach \i in {4,5,6}
     \draw[->] (I3)--(I\i);
 \draw[->] (I1)--(I2);
 \draw[->] (I2)--(I3);
 \draw[->] (I10)--(I3);
 \draw[->] (D1)--(I10);
 
 \node[fit=(I4) (I5) (I6), draw, inner sep=2mm, color=lightgray] (fit1) {};
 \node[fit=(I3) (I10) (D1), draw, inner sep=3mm, color=lightgray] (fit2) {};
 \node[fit=(fit1) (fit2) (I2), draw, inner sep=1mm, color=gray]
     (fit4) {};

 \node[draw=gray, rectangle, fill=white, above right=1.4mm and -16mm of I5] {\footnotesize Framework: PyG};
 \node[draw=gray, rectangle, fill=white, above left=1.4mm and -16mm of I6] {\footnotesize Framework: DGL};
 
 \node[draw=gray, rectangle, fill=white, above=1.4mm of I4] {\footnotesize No Framework Indicated};




\end{tikzpicture}

\caption{Software architecture of gSuite.}
\label{fig-arch}
\end{figure}

gSuite provides an interface that enables researchers and engineers to easily build a desired GNN pipeline in a plug-and-play manner.
We abstract the usage of our benchmark suite from its code implementation to avoid the intervention of end users from coding.
The architecture of underlying software is illustrated in Fig.~\ref{fig-arch}.

When running gSuite, user parameters (e.g. number of layers, GNN model, dataset) are passed to the \textit{User Interface}.
These parameters are interpreted by the interface and then passed to the \textit{Abstraction Module}.
Nevertheless, the interface does not require the end user to pass all the parameters to the suite.
There is a configuration file that includes all these settings as default parameters, where these default parameters take action when a parameter value is not specified by the user.

The decision of which framework, GNN model and dataset are going to be used is made by this abstraction module.
In case of no framework is indicated by the end user; then our GNN implementations are utilized.

\begin{figure}[h]
  \centering
  
\begin{tikzpicture}[
 show background rectangle]

 \node[rectangle, draw, align=center, fill=yellow, font=\footnotesize]
 (I1) {edgeIndex\\ \textcolor{gray}{$(COO)$} };
 
 \node[module, below=3mm of I1, fill=orange, font=\footnotesize, align=center]
 (I2) {scatter};
 
 \node[rectangle, draw, align=center, below=3mm of I2, fill=yellow, font=\footnotesize]
 (I3) {nodeDegrees\\ \textcolor{gray}{$[1~x~n]$} };
 
 \node[rectangle, draw, align=center, right=3mm of I1, fill=yellow, font=\footnotesize]
 (I4) {featureVector\\ \textcolor{gray}{$[n~x~f]$} };
 
 \node[module, below=3mm of I4, fill=orange, font=\footnotesize, align=center]
 (I5) {sgemm};
 
 \node[rectangle, draw, align=center, below=3mm of I5, fill=yellow, font=\footnotesize]
 (I6) {linearOutput\\ \textcolor{gray}{$[n~x~o]$} };
 
 \node[module, below right=4mm and -7mm of I3, fill=orange, font=\footnotesize, align=center]
 (I7) {indexSelect};
 
 \node[rectangle, draw, below right=3mm and -8mm of I7, fill=yellow, font=\footnotesize, align=center]
 (I8) {indexSelect\\Output\\ \textcolor{gray}{$[e~x~o]$} };
 
 \node[module, below left=3mm and -3mm of I8, fill=orange, font=\footnotesize, align=center]
 (I9) {scatter};
 
 \node[rectangle, draw, align=center, left=3mm of I8, fill=yellow, font=\footnotesize]
 (I10) {edgeIndex\\ \textcolor{gray}{$(COO)$} };
 
 \node[rectangle, draw, align=center, below=3mm of I9, fill=yellow, font=\footnotesize]
 (I11) {featureVectorOutput\\ \textcolor{gray}{$[n~x~o]$}};
 
 
 %
 %
 %
 \node[rectangle, above right=2mm and -7mm of I1, align=center, font=\bfseries]
 (TH) {gSuite-MP};
 
 \draw[->] (I1)--(I2);
 \draw[->] (I2)--(I3);
 \draw[->] (I4)--(I5);
 \draw[->] (I5)--(I6);
 \draw[->] (I3)--(I7);
 \draw[->] (I6)--(I7);
 \draw[->] (I7)--(I8);
 \draw[->] (I8)--(I9);
 \draw[->] (I10)--(I9);
 \draw[->] (I9)--(I11);




\end{tikzpicture} 
\begin{tikzpicture}[
 show background rectangle]

 \node[rectangle, draw, align=center, fill=yellow, font=\footnotesize]
 (I1) {$D^{(1/2)}$\\ \textcolor{gray}{$[n~x~n]$}};
 
 \node[rectangle, draw, align=center, right=3mm of I1, fill=yellow, font=\footnotesize]
 (I2) {$A$\\ \textcolor{gray}{$[n~x~n]$}};
 
 \node[module, below right=3mm and -4mm of I1, fill=orange, font=\footnotesize, align=center]
 (I3) {SpGEMM};
 
 \node[rectangle, draw, align=center, below left=3mm and -4mm of I3, fill=yellow, font=\footnotesize]
 (I4) {$D^{(1/2)}*A$\\ \textcolor{gray}{$[n~x~n]$}};
 
 \node[rectangle, right=3mm of I4, draw, align=center,fill=yellow, font=\footnotesize]
 (I5) {$D^{(1/2)}$\\ \textcolor{gray}{$[n~x~n]$}};
 
\node[module, below right=3mm and -6mm of I4, fill=orange, font=\footnotesize, align=center]
 (I6) {SpGEMM};
 
 \node[rectangle, draw, align=center, below left=3mm and -8mm of I6, fill=yellow, font=\footnotesize]
 (I7) {$D^{(1/2)}*A*D^{(1/2)}$\\ \textcolor{gray}{$[n~x~n]$}};
 
 \node[rectangle, draw, align=center, right=3mm of I7, fill=yellow, font=\footnotesize]
 (I8) {$X$\\ \textcolor{gray}{$[n~x~f]$}};
 
 \node[module, below right=3mm and -6mm of I7, fill=orange, font=\footnotesize, align=center]
 (I9) {SpGEMM};
 
 \node[rectangle, draw, align=center, below left=3mm and -13mm of I9, fill=yellow, font=\footnotesize]
 (I10) {$D^{(1/2)}*A*D^{(1/2)}*X$\\ \textcolor{gray}{$[n~x~f]$}};

 %
 %
 %
 %
 \node[rectangle, above right=1mm and -15mm of I1, align=center, font=\bfseries]
 (TH) {gSuite-SpMM};
 
 \draw[->] (I1)--(I3);
 \draw[->] (I2)--(I3);
 \draw[->] (I3)--(I4);
 \draw[->] (I4)--(I6);
 \draw[->] (I5)--(I6);
 \draw[->] (I6)--(I7);
 \draw[->] (I7)--(I9);
 \draw[->] (I8)--(I9);
 \draw[->] (I9)--(I10);




\end{tikzpicture}

\caption{Computational schema of the GCN pipelines of gSuite-MP and gSuite-SpMM. Yellow boxes represent data, orange ones represent the \emph{core kernels}.}
\label{fig-computation-schema}
\end{figure}
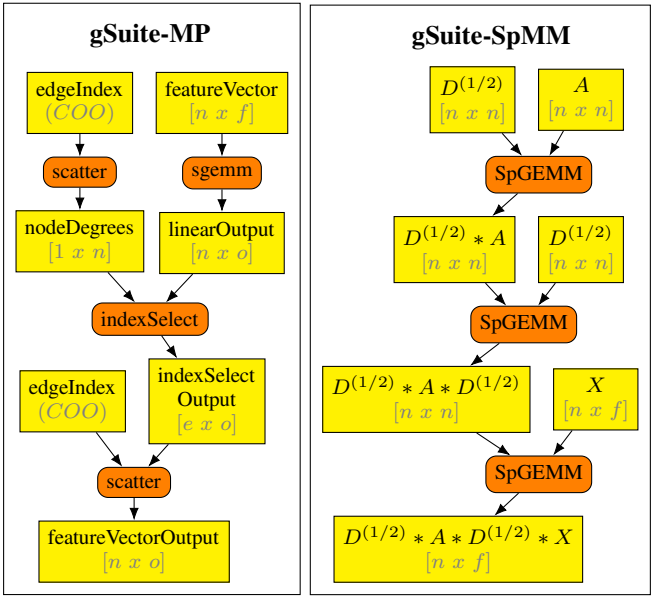

Data loader imports the chosen dataset and handles pre-processing stage of GNN computation, i.e., loads edge index vector and feature representation vector.

gSuite implements both of the computation models (MP and SpMM) for each deployed GNN model.
Iterative execution of these kernels with proper data manipulation results in a GNN model.
To illustrate the phenomenon, graph convolutional network (GCN) inference is implemented by adding \textit{indexSelect}, \textit{scatter} and \textit{linear} kernels to satisfy GCN's MP computation scheme.
On the other hand, the SpMM computation of GCN refers to reducing all the above-mentioned operations in a single matrix multiplication.
We implemented SpMM GCN by utilizing NVIDIA's cuBlas utilities.
An illustration of these implementations are given at Fig.~\ref{fig-computation-schema}.

All the implemented kernels are listed in Table~\ref{tab:kernels} with their brief description.
These kernels are designed to be generic and GNN-oriented so that any GNN model can be built by utilizing these kernels.


\section{Evaluation}\label{evaluation}
In this section, we first explain the GNN models and datasets that we deployed in our suite.
Next, our experimental setup is briefly described.
Finally, we deliver the results of our experiments and discuss our observations.

\subsection{GNN Models}
We implemented the most potent GNN models in our benchmark suite: GCN, GIN and SAG.
We made the model implementations two-sided for computational model versatility, i.e., each model has distinct MP and SpMM implementations (except SAG).

MP models consist of \emph{indexSelect}, \emph{scatter}, and \emph{sgemm} kernels; SpMM models incorporate \emph{SpMM} and \emph{sgemm} kernels.

\subsection{Datasets}
In our evaluation, we used the most prevalent graph datasets across varying domains: Cora, Citeseer\cite{cora}, Pubmed\cite{pubmed}, Reddit\cite{reddit}, and LiveJournal\cite{LJdataset1,LJdataset2}.
These datasets vary greatly in terms of size, feature length, number of nodes and edges.
A table of datasets with their information is given in Table~\ref{tab:datasets}.

\begin{table}
  \caption{The Included Datasets in our Evaluation}
  \label{tab:datasets}
  \begin{tabular}{lcccc}
    \toprule
    \textbf{Dataset}     & \textbf{Nodes}  & \textbf{Feature}    & \textbf{Edges}  & \textbf{Short}\\
                         &                 & \textbf{Length}     &                 & \textbf{Form}\\
    \midrule
    \textbf{Cora}\cite{cora}          & 2,708     & 1,433 & 5,429 & CR\\
    \textbf{CiteSeer}\cite{cora}      & 3,327   & 3,703   & 4,732 & CS\\
    \textbf{PubMed}\cite{pubmed}      & 19,717    & 500   & 44,438 & PB\\
    \textbf{Reddit}\cite{reddit}    & 232,965 & 602 & 11,606,919 & RD\\
    \textbf{LiveJournal}\cite{LJdataset1}    & 	4,847,571 & 1 & 68,993,773 & LJ\\
  \bottomrule
\end{tabular}
\end{table}

Each of these datasets represents a particular feature that aims to test the limitations of implemented GNN models and underlying architectures.
Each attribute of the datasets makes them unique in terms of computation. For instance, one may include a large amount of feature size; while other may have a huge number of directed edges.


\subsection{Experimental Setup}
Our experiments were conducted on NVIDIA V100 GPU 32GB and Intel Xeon 2000 CPU. Each GNN model with the specified input set is run three times; and the mean values of the statistics of these runs were collected. Profiling operations are done at the kernel level for all GNN pipelines.

NVIDIA's nvprof\cite{nvprofiler} is used for collecting the results on the GPU card.
It is a profiling tool that tracks running applications on GPUs and collects information about the performance activity of the application.
We use the version 10.2 of nvprof.

GPGPU-Sim\cite{gpgpu-sim} is utilized for collecting more detailed performance statistics.
GPGPU-Sim is a timing detailed architectural simulator which is capable of running CUDA and OpenCL kernels.
We use the configuration file that is provided by the simulator package and models a GPU architecture similar to NVIDIA's V100 GPU.
We used the version 4.0 of GPGPU-Sim.



\subsection{Results}
%
%
%

\subsubsection{Execution Time}
\begin{figure}[h]
  \centering

\begin{tikzpicture}[framed]



\pgfplotsset{every tick label/.append style={font=\footnotesize}}

  \begin{axis}[width=3cm,
    align=center,
    ybar,
    axis x line = none,
    y axis line style = { opacity = 0 },
    ymin=0,
    ymax=4000,
    tickwidth = 0pt,
    xtick={1},
    ytick={0, 2000, 4000},
    yticklabels = {0,2,4},
    ylabel style = {font=\footnotesize},
    ]
    \addplot [draw = white,
    semithick,
    pattern = crosshatch,
    fill = white]
    coordinates { (1,90) };
  \end{axis}
  
  \begin{axis}[width=3cm,
    align=center,
    ybar,
    axis x line = none,
    y axis line style = { opacity = 0 },
    ymin=0,
    ymax=4000,
    yshift=-1.41cm,
    tickwidth = 0pt,
    xtick={1},
    ytick={0, 2000, 4000},
    yticklabels = {0,2,4},
    ylabel style = {font=\footnotesize},
    ]
    \addplot [draw = white,
    semithick,
    pattern = crosshatch,
    fill = white]
    coordinates { (1,90) };
  \end{axis}
  
  \begin{axis}[width=3cm,
    align=center,
    ybar,
    axis x line = none,
    y axis line style = { opacity = 0 },
    ymin=0,
    ymax=4000,
    yshift=-2.73cm,
    tickwidth = 0pt,
    xtick={1},
    ytick={0, 2000, 4000},
    yticklabels = {0,2,4},
    ylabel style = {font=\footnotesize},
    ]
    \addplot [draw = white,
    semithick,
    pattern = crosshatch,
    fill = white]
    coordinates { (1,90) };
  \end{axis}
  
  
  \draw[step=0.25cm,lightgray,very thin, dashed] (0,0) grid (7,1);
  
  \draw[step=0.25cm,lightgray,very thin, dashed] (0,-0.2) grid (7,-1.41);
  
  \draw[step=0.25cm,lightgray,very thin, dashed] (0,-1.7) grid (7,-2.73);

  \node[align=center] at (-0.6,-1) {\rotatebox{90}{\footnotesize{Execution Time (sec)}}};
  \node at (3,-3.5) {\footnotesize Datasets};

  \begin{axis}[width=3cm,
    xshift=-0cm,
    align=center,
    ybar,
    bar width=0.2cm,
    x axis line style = { opacity = 0 },
    axis y line       = none,
    tickwidth         = 0pt,
    ymin = 0,
    ymax = 4000,
    xmin=CR,
    xmax=CR,
    xtick={CR},
    xticklabel style={rotate=90},
    symbolic x coords = {CR}
  ]
  \addplot [draw = blue0,
    semithick,
    pattern = crosshatch,
    fill = blue0]
    coordinates { (CR,3210) };
  \addplot [draw = blue2,
    semithick,
    pattern = north west lines,
    fill = blue2]
    coordinates { (CR,767) };
  \addplot [draw = blue1,
    semithick,
    pattern = north west lines,
    pattern color = blue1]
    coordinates { (CR,771) };
  \addplot [draw = blue3,
    semithick,
    pattern = crosshatch,
    pattern color = blue3]
    coordinates { (CR,881) };
  \end{axis}
  
  \begin{axis}[width=3cm,
    xshift=1.3cm,
    align=center,
    title style = {font=\footnotesize},
    ybar,
    bar width=0.2cm,
    x axis line style = { opacity = 0 },
    axis y line       = none,
    tickwidth         = 0pt,
    ymin = 0,
    ymax = 4000,
    xmin=CS,
    xmax=CS,
    xtick={CS},
    xticklabel style={rotate=90},
    symbolic x coords = {CS},
  ]
  \addplot [draw = blue0,
    semithick,
    pattern = crosshatch,
    fill = blue0]
    coordinates { (CS,3225) };
  \addplot [draw = blue2,
    semithick,
    pattern = north west lines,
    fill = blue2]
    coordinates { (CS,1022) };
  \addplot [draw = blue1,
    semithick,
    pattern = north west lines,
    pattern color = blue1]
    coordinates { (CS,809) };
  \addplot [draw = blue3,
    semithick,
    pattern = crosshatch,
    pattern color = blue3]
    coordinates { (CS,1043) };
  \end{axis}

  \begin{axis}[legend style={at={(-1.8,-3.8)},anchor=north west,nodes={scale=0.7, transform shape},
    /tikz/every even column/.append style={column sep=0.2cm}},
    legend columns=4,
    align=center,
    width=3cm,
    xshift=2.6cm,
    ybar,
    bar width=0.2cm,
    x axis line style = { opacity = 0 },
    axis y line       = none,
    tickwidth         = 0pt,
    ymin = 0,
    ymax = 4000,
    xmin=PB,
    xmax=PB,
    xtick={PB},
    xticklabel style={rotate=90},
    symbolic x coords = {PB},
  ]
  \addplot [draw = blue0,
    semithick,
    pattern = crosshatch,
    fill = blue0]
    coordinates { (PB,3219) };
  \addplot [draw = blue2,
    semithick,
    pattern = north west lines,
    fill = blue2]
    coordinates { (PB,2564) };
  \addplot [draw = blue1,
    semithick,
    pattern = north west lines,
    pattern color = blue1]
    coordinates { (PB,802) };
  \addplot [draw = blue3,
    semithick,
    pattern = crosshatch,
    pattern color = blue3]
    coordinates { (PB,1495) };
  \legend{PyG, DGL, gSuite-MP, gSuite-SpMM}
  \end{axis}

    \begin{axis}[width=3cm,
    xshift=3.9cm,
    align=center,
    ybar,
    bar width=0.2cm,
    x axis line style = { opacity = 0 },
    axis y line       = none,
    tickwidth         = 0pt,
    ymin = 0,
    ymax = 4000,
    xmin=RD,
    xmax=RD,
    xtick={RD},
    xticklabel style={rotate=90},
    symbolic x coords = {RD}
  ]
  \addplot [draw = blue0,
    semithick,
    pattern = crosshatch,
    fill = blue0]
    coordinates { (RD,3410) };
  \addplot [draw = blue2,
    semithick,
    pattern = north west lines,
    fill = blue2]
    coordinates { (RD,950) };
  \addplot [draw = blue1,
    semithick,
    pattern = north west lines,
    pattern color = blue1]
    coordinates { (RD,820) };
  \addplot [draw = blue3,
    semithick,
    pattern = crosshatch,
    pattern color = blue3]
    coordinates { (RD,1020) };
  \end{axis}

  \begin{axis}[legend style={at={(0.9,1)},anchor=north west},
    align=center,
    width=3cm,
    xshift=5.2cm,
    ybar,
    bar width=0.2cm,
    x axis line style = { opacity = 0 },
    axis y line       = none,
    tickwidth         = 0pt,
    ymin = 0,
    ymax = 4000,
    xmin=LJ,
    xmax=LJ,
    xtick={LJ},
    xticklabel style={rotate=90},
    symbolic x coords = {LJ},
  ]
  \addplot [draw = blue0,
    semithick,
    pattern = crosshatch,
    fill = blue0]
    coordinates { (LJ,3615) };
  \addplot [draw = blue2,
    semithick,
    pattern = north west lines,
    fill = blue2]
    coordinates { (LJ,2495) };
  \addplot [draw = blue1,
    semithick,
    pattern = north west lines,
    pattern color = blue1]
    coordinates { (LJ,1135) };
  \addplot [draw = blue3,
    semithick,
    pattern = crosshatch,
    pattern color = blue3]
    coordinates { (LJ,1270) };
  \end{axis}
  
  

  \begin{axis}[width=3cm,
    align=center,
    xshift=-0cm,
    yshift=-1.4cm,
    ybar,
    bar width=0.2cm,
    x axis line style = { opacity = 0 },
    axis y line       = none,
    tickwidth         = 0pt,
    ymin = 0,
    ymax = 4000,
    xmin=CR,
    xmax=CR,
    xtick={CR},
    xticklabel style={rotate=90},
    symbolic x coords = {CR},
  ]
  \addplot [draw = blue0,
    semithick,
    pattern = crosshatch,
    fill = blue0]
    coordinates { (CR,3552) };
  \addplot [draw = blue2,
    semithick,
    pattern = north west lines,
    fill = blue2]
    coordinates { (CR,770) };
  \addplot [draw = blue1,
    semithick,
    pattern = north west lines,
    pattern color = blue1]
    coordinates { (CR,912) };
  \addplot [draw = blue3,
    semithick,
    pattern = crosshatch,
    pattern color = blue3]
    coordinates { (CR,595) };
  \end{axis}
  
  \begin{axis}[width=3cm,
    xshift=1.3cm,
    yshift=-1.4cm,
    align=center,
    title style = {font=\footnotesize},
    ybar,
    bar width=0.2cm,
    x axis line style = { opacity = 0 },
    axis y line       = none,
    tickwidth         = 0pt,
    ymin = 0,
    ymax = 4000,
    xmin=CS,
    xmax=CS,
    xtick={CS},
    xticklabel style={rotate=90},
    symbolic x coords = {CS},
  ]
  \addplot [draw = blue0,
    semithick,
    pattern = crosshatch,
    fill = blue0]
    coordinates { (CS,3654) };
  \addplot [draw = blue2,
    semithick,
    pattern = north west lines,
    fill = blue2]
    coordinates { (CS,771) };
  \addplot [draw = blue1,
    semithick,
    pattern = north west lines,
    pattern color = blue1]
    coordinates { (CS,1096) };
  \addplot [draw = blue3,
    semithick,
    pattern = crosshatch,
    pattern color = blue3]
    coordinates { (CS,601) };
  \end{axis}
  
  \begin{axis}[legend style={at={(0.9,1)},anchor=north west},
    align=center,
    width=3cm,
    xshift=2.6cm,
    yshift=-1.4cm,
    ybar,
    bar width=0.2cm,
    x axis line style = { opacity = 0 },
    axis y line       = none,
    tickwidth         = 0pt,
    ymin = 0,
    ymax = 4000,
    xmin=PB,
    xmax=PB,
    xtick={PB},
    xticklabel style={rotate=90},
    symbolic x coords = {PB},
  ]
  \addplot [draw = blue0,
    semithick,
    pattern = crosshatch,
    fill = blue0]
    coordinates { (PB,3615) };
  \addplot [draw = blue2,
    semithick,
    pattern = north west lines,
    fill = blue2]
    coordinates { (PB,2495) };
  \addplot [draw = blue1,
    semithick,
    pattern = north west lines,
    pattern color = blue1]
    coordinates { (PB,1135) };
  \addplot [draw = blue3,
    semithick,
    pattern = crosshatch,
    pattern color = blue3]
    coordinates { (PB,774) };
  \end{axis}

  \begin{axis}[legend style={at={(0.9,1)},anchor=north west},
    align=center,
    width=3cm,
    xshift=3.9cm,
    yshift=-1.4cm,
    ybar,
    bar width=0.2cm,
    x axis line style = { opacity = 0 },
    axis y line       = none,
    tickwidth         = 0pt,
    ymin = 0,
    ymax = 4000,
    xmin=RD,
    xmax=RD,
    xtick={RD},
    xticklabel style={rotate=90},
    symbolic x coords = {RD},
  ]
  \addplot [draw = blue0,
    semithick,
    pattern = crosshatch,
    fill = blue0]
    coordinates { (RD,3815) };
  \addplot [draw = blue2,
    semithick,
    pattern = north west lines,
    fill = blue2]
    coordinates { (RD,2795) };
  \addplot [draw = blue1,
    semithick,
    pattern = north west lines,
    pattern color = blue1]
    coordinates { (RD,1435) };
  \addplot [draw = blue3,
    semithick,
    pattern = crosshatch,
    pattern color = blue3]
    coordinates { (RD,1173) };
  \end{axis}

    \begin{axis}[legend style={at={(0.9,1)},anchor=north west},
    align=center,
    width=3cm,
    xshift=5.2cm,
    yshift=-1.4cm,
    ybar,
    bar width=0.2cm,
    x axis line style = { opacity = 0 },
    axis y line       = none,
    tickwidth         = 0pt,
    ymin = 0,
    ymax = 4000,
    xmin=LJ,
    xmax=LJ,
    xtick={LJ},
    xticklabel style={rotate=90},
    symbolic x coords = {LJ},
  ]
  \addplot [draw = blue0,
    semithick,
    pattern = crosshatch,
    fill = blue0]
    coordinates { (LJ,3915) };
  \addplot [draw = blue2,
    semithick,
    pattern = north west lines,
    fill = blue2]
    coordinates { (LJ,2995) };
  \addplot [draw = blue1,
    semithick,
    pattern = north west lines,
    pattern color = blue1]
    coordinates { (LJ,1535) };
  \addplot [draw = blue3,
    semithick,
    pattern = crosshatch,
    pattern color = blue3]
    coordinates { (LJ,1376) };
  \end{axis}



  
  \begin{axis}[width=3cm,
    align=center,
    xshift=-0.14cm,
    yshift=-2.7cm,
    ybar,
    bar width=0.2cm,
    x axis line style = { opacity = 0 },
    axis y line       = none,
    tickwidth         = 0pt,
    ymin = 0,
    ymax = 4000,
    xmin=CR,
    xmax=CR,
    xtick={CR},
    xticklabel style={rotate=90},
    symbolic x coords = {CR},
  ]
  \addplot [draw = blue0,
    semithick,
    pattern = crosshatch,
    fill = blue0]
    coordinates { (CR,3204) };
  \addplot [draw = blue2,
    semithick,
    pattern = north west lines,
    fill = blue2]
    coordinates { (CR,783) };
  \addplot [draw = blue1,
    semithick,
    pattern = north west lines,
    pattern color = blue1]
    coordinates { (CR,906) };
  \end{axis}
  
  \begin{axis}[width=3cm,
    xshift=1.2cm,
    yshift=-2.7cm,
    align=center,
    title style = {font=\footnotesize},
    ybar,
    bar width=0.2cm,
    x axis line style = { opacity = 0 },
    axis y line       = none,
    tickwidth         = 0pt,
    ymin = 0,
    ymax = 4000,
    xmin=CS,
    xmax=CS,
    xtick={CS},
    xticklabel style={rotate=90},
    symbolic x coords = {CS},
  ]
  \addplot [draw = blue0,
    semithick,
    pattern = crosshatch,
    fill = blue0]
    coordinates { (CS,3215) };
  \addplot [draw = blue2,
    semithick,
    pattern = north west lines,
    fill = blue2]
    coordinates { (CS,822) };
  \addplot [draw = blue1,
    semithick,
    pattern = north west lines,
    pattern color = blue1]
    coordinates { (CS,1059) };
  \end{axis}
  
  \begin{axis}[legend style={at={(0.6,2.7)},anchor=north west},
    align=center,
    width=3cm,
    xshift=2.46cm,
    yshift=-2.7cm,
    ybar,
    bar width=0.2cm,
    x axis line style = { opacity = 0 },
    axis y line       = none,
    tickwidth         = 0pt,
    ymin = 0,
    ymax = 4000,
    xmin=PB,
    xmax=PB,
    xtick={PB},
    xticklabel style={rotate=90},
    symbolic x coords = {PB},
  ]
  \addplot [draw = blue0,
    semithick,
    pattern = crosshatch,
    fill = blue0]
    coordinates { (PB,3113) };
  \addplot [draw = blue2,
    semithick,
    pattern = north west lines,
    fill = blue2]
    coordinates { (PB,2537) };
  \addplot [draw = blue1,
    semithick,
    pattern = north west lines,
    pattern color = blue1]
    coordinates { (PB,1078) };
  \end{axis}

  \begin{axis}[legend style={at={(0.6,2.7)},anchor=north west},
    align=center,
    width=3cm,
    xshift=3.8cm,
    yshift=-2.7cm,
    ybar,
    bar width=0.2cm,
    x axis line style = { opacity = 0 },
    axis y line       = none,
    tickwidth         = 0pt,
    ymin = 0,
    ymax = 4000,
    xmin=RD,
    xmax=RD,
    xtick={RD},
    xticklabel style={rotate=90},
    symbolic x coords = {RD},
  ]
  \addplot [draw = blue0,
    semithick,
    pattern = crosshatch,
    fill = blue0]
    coordinates { (RD,3413) };
  \addplot [draw = blue2,
    semithick,
    pattern = north west lines,
    fill = blue2]
    coordinates { (RD,2937) };
  \addplot [draw = blue1,
    semithick,
    pattern = north west lines,
    pattern color = blue1]
    coordinates { (RD,1678) };
  \end{axis}

  \begin{axis}[legend style={at={(0.6,2.7)},anchor=north west},
    align=center,
    width=3cm,
    xshift=5.06cm,
    yshift=-2.7cm,
    ybar,
    bar width=0.2cm,
    x axis line style = { opacity = 0 },
    axis y line       = none,
    tickwidth         = 0pt,
    ymin = 0,
    ymax = 4000,
    xmin=LJ,
    xmax=LJ,
    xtick={LJ},
    xticklabel style={rotate=90},
    symbolic x coords = {LJ},
  ]
  \addplot [draw = blue0,
    semithick,
    pattern = crosshatch,
    fill = blue0]
    coordinates { (LJ,3813) };
  \addplot [draw = blue2,
    semithick,
    pattern = north west lines,
    fill = blue2]
    coordinates { (LJ,3137) };
  \addplot [draw = blue1,
    semithick,
    pattern = north west lines,
    pattern color = blue1]
    coordinates { (LJ,1978) };
  \end{axis}


\end{tikzpicture}
\caption{End-to-end execution time of frameworks with different GNN models on varying datasets.}
\label{fig:end-to-end-exec-time}
\end{figure}
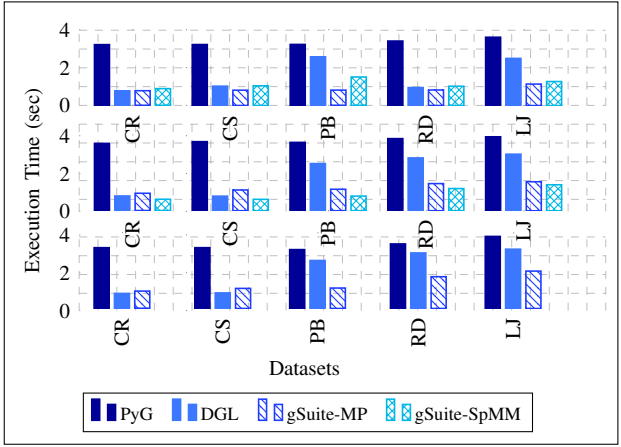

We start our evaluation by measuring execution time of GNN pipelines; using their implementations with PyG, DGL, gSuite-MP, and gSuite-SpMM.
We measure the execution time as wall clock time.
In general, execution times of PyG is longer than other frameworks.
This is mainly because of the initializations that are performed as part of their  implementation.
Since the gSuite eliminates high level library dependencies, its implementations tend to run faster than other frameworks in terms of end-to-end execution time.
We compare these durations in Fig.~\ref{fig:end-to-end-exec-time}.

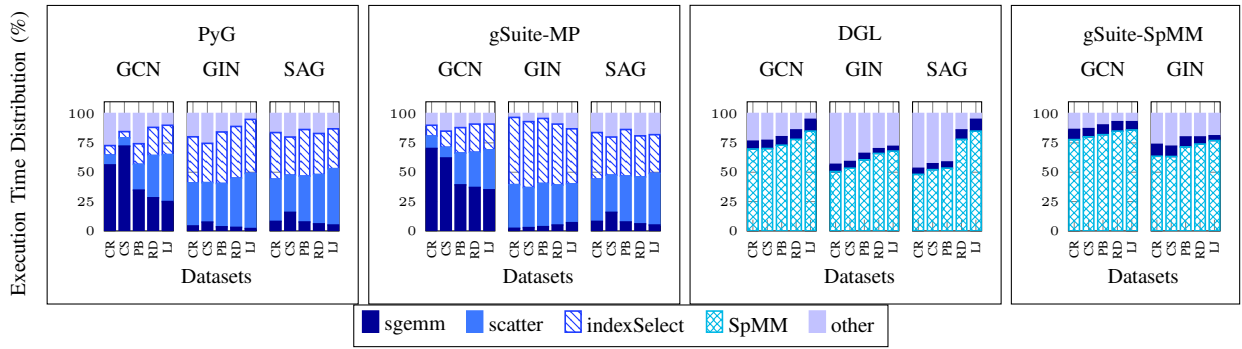
\begin{figure*}[h]
  \centering
  
  \pgfplotsset{every tick label/.append style={font=\tiny}}

\pgfplotstableread{ 
Label       sgemm   scatter  indexSelect    other
Cora        56.2    8.4      8.1            27.3
CiteSeer    72.2    7.1      5.3            15.4
Pubmed      34.7    22       17.6           25.7
Reddit      28.2    36       24             11.8
LiveJournal 25      40       25             10
    }\datapyggcn
    
\pgfplotstableread{ 
Label       sgemm   scatter  indexSelect    other
Cora        4.2     36.7     39.3           19.8
CiteSeer    7.5     33.6     33.5           25.4
Pubmed      3.5     36.9     43.9           15.7
Reddit      3       42       44             11
LiveJournal 2       47       46             5
    }\datapyggin
    
\pgfplotstableread{ 
Label       sgemm   scatter  indexSelect    other
Cora        8.1     36.0     39.7           16.2
CiteSeer    15.8    31.7     32.5           20
Pubmed      7.6     39.1     39.7           13.6
Reddit      6       42       35             17
LiveJournal 5       48       34             13
    }\datapygsag

\pgfplotstableread{ 
Label       SpMM    sgemm    other
Cora        69.8    6.6      23.6
CiteSeer    70.2    6.9      22.9
Pubmed      73.0    7.2      19.8
Reddit      78      8        14
LiveJournal 85      10       5
    }\datadglgcn
    
\pgfplotstableread{ 
Label       SpMM    sgemm    other
Cora        50.6    6.1      43.3
CiteSeer    53.4    5.7      40.9
Pubmed      60.6    5.4      34
Reddit      66      4        30
LiveJournal 68      4        28
    }\datadglgin
    
\pgfplotstableread{ 
Label       SpMM    sgemm    other
Cora        48.2    5.1      46.7
CiteSeer    52.2    5.0      42.8
Pubmed      53.5    5.2      41.3
Reddit      78      8        14
LiveJournal 85      10       5
    }\datadglsag

\pgfplotstableread{ 
Label       sgemm   scatter  indexSelect    other
Cora        70.2    10.6     9.2            10
CiteSeer    62.1    9.1      13.8           15
Pubmed      39.2    27.1     21.7           12
Reddit      37      30       24             9
LiveJournal 35      34       22             9
    }\datagsuitempgcn
    
\pgfplotstableread{ 
Label       sgemm   scatter  indexSelect    other
Cora        2.2     37.2     57.3           3.3
CiteSeer    2.8     34.1     56.2           6.9
Pubmed      3.6     36.8     55.4           4.2
Reddit      5       34       52             9
LiveJournal 7       33       47             13
    }\datagsuitempgin
    
\pgfplotstableread{ 
Label       sgemm   scatter  indexSelect    other
Cora        8.1     36.0     39.7           16.2
CiteSeer    15.8    31.7     32.5           20
Pubmed      7.6     39.1     39.7           13.6
Reddit      6       40       35             19
LiveJournal 5       44       33             18
    }\datagsuitempsag

\pgfplotstableread{ 
Label       SpMM    sgemm    other
Cora        77.8    8.6      13.6
CiteSeer    80.2    6.9      12.9
Pubmed      82.0    8.2      9.8
Reddit      85      8        7
LiveJournal 86      7        7
    }\datagsuitespmmgcn
    
\pgfplotstableread{ 
Label       SpMM    sgemm    other
Cora        63.6    10.1     26.3
CiteSeer    63.4    8.7      27.9
Pubmed      71.6    8.4      20
Reddit      74      6        20
LiveJournal 77      4        19
    }\datagsuitespmmgin
    
\pgfplotstableread{ 
Label       SpMM    sgemm    other
Cora        62.2    8.1      29.7
CiteSeer    65.2    9.0      26.8
Pubmed      63.5    10.2     26.3
Reddit      65      12       23
LiveJournal 68      15       17
    }\datagsuitespmmsag
    
\pgfplotstableread{ 
Label       sgemm   scatter  indexSelect   SpMM  other
Cora        0       0        0             0     0
}\datalegend

    \begin{tikzpicture}
        \node[align=center] at (0,2) {\rotatebox{90}{\footnotesize{Execution Time Distribution (\%)}}};
    \end{tikzpicture}
    \begin{tikzpicture}[framed]

    \node at (1.5,2.6) {\footnotesize PyG};
    \node at (1.5,-0.6) {\footnotesize Datasets};

    \begin{axis}[
    width=2.5cm,
    height=3.3cm,
    title={GCN},
    title style = {font=\footnotesize},
    ybar stacked,   
    bar width = 0.14cm,
    ymin=0,         
    xtick=data,     
    legend style={at={(axis cs:105,0)},anchor=south west, font=\footnotesize},
    ytick={0,25,50,75,100},
    xticklabels = {CR,CS,PB,RD,LJ},  
    xticklabel style={rotate=90}
    ]
    \addplot [draw = blue0,
    semithick,
    pattern = north west lines,
    fill = blue0] table [y=sgemm, meta=Label,x expr=\coordindex] {\datapyggcn};   
    \addplot [draw = blue2,
    semithick,
    pattern = crosshatch,
    fill = blue2] table [y=scatter, meta=Label,x expr=\coordindex] {\datapyggcn};
    \addplot [draw = blue1,
    semithick,
    pattern = north west lines,
    pattern color = blue1] table [y=indexSelect, meta=Label,x expr=\coordindex] {\datapyggcn};
    \addplot [draw = blue4,
    semithick,
    pattern = grid,
    fill = blue4] table [y=other, meta=Label,x expr=\coordindex] {\datapyggcn};
    \end{axis}
    
    \begin{axis}[
    width=2.5cm,
    height=3.3cm,
    title={GIN},
    title style = {font=\footnotesize},
    xshift=1.1cm,
    ybar stacked,   
    bar width = 0.14cm,
    ymin=0,         
    xtick=data,     
    legend style={at={(axis cs:8,-7.5)},anchor=south west},
    yticklabels={,,},
    xticklabels = {CR,CS,PB,RD,LJ},
    xticklabel style={rotate=90}
    ]
    \addplot [draw = blue0,
    semithick,
    pattern = north west lines,
    fill = blue0] table [y=sgemm, meta=Label,x expr=\coordindex] {\datapyggin};   
    \addplot [draw = blue2,
    semithick,
    pattern = crosshatch,
    fill = blue2] table [y=scatter, meta=Label,x expr=\coordindex] {\datapyggin};
    \addplot [draw = blue1,
    semithick,
    pattern = north west lines,
    pattern color = blue1] table [y=indexSelect, meta=Label,x expr=\coordindex] {\datapyggin};
    \addplot [draw = blue4,
    semithick,
    pattern = grid,
    fill = blue4] table [y=other, meta=Label,x expr=\coordindex] {\datapyggin};
    \end{axis}
    
    \begin{axis}[
    width=2.5cm,
    height=3.3cm,
    xshift=2.2cm,
    title={SAG},
    title style = {font=\footnotesize},
    ybar stacked,   
    bar width = 0.14cm,
    ymin=0,         
    xtick=data,     
    legend style={at={(axis cs:55,0)},anchor=south west, font=\footnotesize},
    yticklabels={,,},
    xticklabels = {CR,CS,PB,RD,LJ},  
    xticklabel style={rotate=90}
    ]
    \addplot [draw = blue0,
    semithick,
    pattern = north west lines,
    fill = blue0] table [y=sgemm, meta=Label,x expr=\coordindex] {\datapygsag};   
    \addplot [draw = blue2,
    semithick,
    pattern = crosshatch,
    fill = blue2] table [y=scatter, meta=Label,x expr=\coordindex] {\datapygsag};
    \addplot [draw = blue1,
    semithick,
    pattern = north west lines,
    pattern color = blue1] table [y=indexSelect, meta=Label,x expr=\coordindex] {\datapygsag};
    \addplot [draw = blue4,
    semithick,
    pattern = grid,
    fill = blue4] table [y=other, meta=Label,x expr=\coordindex] {\datapygsag};
    \end{axis}

    \end{tikzpicture}
    \begin{tikzpicture}[framed]

    \node at (1.45,2.6) {\footnotesize gSuite-MP};
    \node at (1.5,-0.6) {\footnotesize Datasets};

    \begin{axis}[
    width=2.5cm,
    height=3.3cm,
    title={GCN},
    title style = {font=\footnotesize},
    ybar stacked,   
    bar width = 0.14cm,
    ymin=0,         
    xtick=data,     
    legend style={at={(axis cs:105,0)},anchor=south west, font=\footnotesize},
    ytick={0,25,50,75,100},
    xticklabels = {CR,CS,PB,RD,LJ},  
    xticklabel style={rotate=90}
    ]
    \addplot [draw = blue0,
    semithick,
    pattern = north west lines,
    fill = blue0] table [y=sgemm, meta=Label,x expr=\coordindex] {\datagsuitempgcn};   
    \addplot [draw = blue2,
    semithick,
    pattern = crosshatch,
    fill = blue2] table [y=scatter, meta=Label,x expr=\coordindex] {\datagsuitempgcn};
    \addplot [draw = blue1,
    semithick,
    pattern = north west lines,
    pattern color = blue1] table [y=indexSelect, meta=Label,x expr=\coordindex] {\datagsuitempgcn};
    \addplot [draw = blue4,
    semithick,
    pattern = grid,
    fill = blue4] table [y=other, meta=Label,x expr=\coordindex] {\datagsuitempgcn};
    \end{axis}
    
    \begin{axis}[
    width=2.5cm,
    height=3.3cm,
    title={GIN},
    title style = {font=\footnotesize},
    xshift=1.1cm,
    ybar stacked,   
    bar width = 0.14cm,
    ymin=0,         
    xtick=data,     
    legend style={at={(axis cs:8,-7.5)},anchor=south west},
    yticklabels={,,},
    xticklabels = {CR,CS,PB,RD,LJ},
    xticklabel style={rotate=90}
    ]
    \addplot [draw = blue0,
    semithick,
    pattern = north west lines,
    fill = blue0] table [y=sgemm, meta=Label,x expr=\coordindex] {\datagsuitempgin};   
    \addplot [draw = blue2,
    semithick,
    pattern = crosshatch,
    fill = blue2] table [y=scatter, meta=Label,x expr=\coordindex] {\datagsuitempgin};
    \addplot [draw = blue1,
    semithick,
    pattern = north west lines,
    pattern color = blue1] table [y=indexSelect, meta=Label,x expr=\coordindex] {\datagsuitempgin};
    \addplot [draw = blue4,
    semithick,
    pattern = grid,
    fill = blue4] table [y=other, meta=Label,x expr=\coordindex] {\datagsuitempgin};
    \end{axis}
    
    \begin{axis}[
    width=2.5cm,
    height=3.3cm,
    xshift=2.2cm,
    title={SAG},
    title style = {font=\footnotesize},
    ybar stacked,   
    bar width = 0.14cm,
    ymin=0,         
    xtick=data,     
    legend style={at={(axis cs:3,0)},anchor=south west, font=\footnotesize},
    yticklabels={,,},
    xticklabels = {CR,CS,PB,RD,LJ},  
    xticklabel style={rotate=90}
    ]
    \addplot [draw = blue0,
    semithick,
    pattern = north west lines,
    fill = blue0] table [y=sgemm, meta=Label,x expr=\coordindex] {\datagsuitempsag};   
    \addplot [draw = blue2,
    semithick,
    pattern = crosshatch,
    fill = blue2] table [y=scatter, meta=Label,x expr=\coordindex] {\datagsuitempsag};
    \addplot [draw = blue1,
    semithick,
    pattern = north west lines,
    pattern color = blue1] table [y=indexSelect, meta=Label,x expr=\coordindex] {\datagsuitempsag};
    \addplot [draw = blue4,
    semithick,
    pattern = grid,
    fill = blue4] table [y=other, meta=Label,x expr=\coordindex] {\datagsuitempsag};
    \end{axis}

    \end{tikzpicture}
    \begin{tikzpicture}[framed]

    \node at (1.5,2.63) {\footnotesize DGL};
    \node at (1.5,-0.6) {\footnotesize Datasets};

    \begin{axis}[
    width=2.5cm,
    height=3.3cm,
    title={GCN},
    title style = {font=\footnotesize},
    ybar stacked,   
    bar width = 0.14cm,
    ymin=0,         
    xtick=data,     
    legend style={at={(axis cs:105,0)},anchor=south west, font=\footnotesize},
    ytick={0,25,50,75,100},
    xticklabels = {CR,CS,PB,RD,LJ},  
    xticklabel style={rotate=90}
    ]
    \addplot [draw = blue3,
    semithick,
    pattern = crosshatch,
    pattern color=blue3] table [y=SpMM, meta=Label,x expr=\coordindex] {\datadglgcn};   
    \addplot [draw = blue0,
    semithick,
    pattern = north west lines,
    fill = blue0] table [y=sgemm, meta=Label,x expr=\coordindex] {\datadglgcn};
    \addplot [draw = blue4,
    semithick,
    pattern = grid,
    fill = blue4] table [y=other, meta=Label,x expr=\coordindex] {\datadglgcn};
    \end{axis}
    
    \begin{axis}[
    width=2.5cm,
    height=3.3cm,
    title={GIN},
    title style = {font=\footnotesize},
    xshift=1.1cm,
    ybar stacked,   
    bar width = 0.14cm,
    ymin=0,         
    xtick=data,     
    legend style={at={(axis cs:8,-7.5)},anchor=south west},
    yticklabels={,,},
    xticklabels = {CR,CS,PB,RD,LJ},
    xticklabel style={rotate=90}
    ]
    \addplot [draw = blue3,
    semithick,
    pattern = crosshatch,
    pattern color=blue3] table [y=SpMM, meta=Label,x expr=\coordindex] {\datadglgin};   
    \addplot [draw = blue0,
    semithick,
    pattern = north west lines,
    fill = blue0] table [y=sgemm, meta=Label,x expr=\coordindex] {\datadglgin};
    \addplot [draw = blue4,
    semithick,
    pattern = grid,
    fill = blue4] table [y=other, meta=Label,x expr=\coordindex] {\datadglgin};
    \end{axis}
    
    \begin{axis}[
    width=2.5cm,
    height=3.3cm,
    xshift=2.2cm,
    title={SAG},
    title style = {font=\footnotesize},
    ybar stacked,   
    bar width = 0.14cm,
    ymin=0,         
    xtick=data,     
    legend style={at={(axis cs:55,0)},anchor=south west, font=\footnotesize},
    yticklabels={,,},
    xticklabels = {CR,CS,PB,RD,LJ},  
    xticklabel style={rotate=90}
    ]
    \addplot [draw = blue3,
    semithick,
    pattern = crosshatch,
    pattern color=blue3] table [y=SpMM, meta=Label,x expr=\coordindex] {\datadglsag};   
    \addplot [draw = blue0,
    semithick,
    pattern = north west lines,
    fill = blue0] table [y=sgemm, meta=Label,x expr=\coordindex] {\datadglsag};
    \addplot [draw = blue4,
    semithick,
    pattern = grid,
    fill = blue4] table [y=other, meta=Label,x expr=\coordindex] {\datadglsag};
    \end{axis}

    \end{tikzpicture}
    \begin{tikzpicture}[framed]

    \node at (1,2.6) {\footnotesize gSuite-SpMM};
    \node at (1,-0.6) {\footnotesize Datasets};

    \begin{axis}[
    width=2.5cm,
    height=3.3cm,
    title={GCN},
    title style = {font=\footnotesize},
    ybar stacked,   
    bar width = 0.14cm,
    ymin=0,         
    xtick=data,     
    legend style={at={(axis cs:105,0)},anchor=south west, font=\footnotesize},
    ytick={0,25,50,75,100},
    xticklabels = {CR,CS,PB,RD,LJ},  
    xticklabel style={rotate=90}
    ]
    \addplot [draw = blue3,
    semithick,
    pattern = crosshatch,
    pattern color=blue3] table [y=SpMM, meta=Label,x expr=\coordindex] {\datagsuitespmmgcn};   
    \addplot [draw = blue0,
    semithick,
    pattern = north west lines,
    fill = blue0] table [y=sgemm, meta=Label,x expr=\coordindex] {\datagsuitespmmgcn};
    \addplot [draw = blue4,
    semithick,
    pattern = grid,
    fill = blue4] table [y=other, meta=Label,x expr=\coordindex] {\datagsuitespmmgcn};
    \end{axis}
    
    \begin{axis}[
    width=2.5cm,
    height=3.3cm,
    title={GIN},
    title style = {font=\footnotesize},
    xshift=1.1cm,
    ybar stacked,   
    bar width = 0.14cm,
    ymin=0,         
    xtick=data,     
    legend style={at={(axis cs:8,-7.5)},anchor=south west},
    yticklabels={,,},
    xticklabels = {CR,CS,PB,RD,LJ},
    xticklabel style={rotate=90}
    ]
     \addplot [draw = blue3,
    semithick,
    pattern = crosshatch,
    pattern color=blue3] table [y=SpMM, meta=Label,x expr=\coordindex] {\datagsuitespmmgin};   
    \addplot [draw = blue0,
    semithick,
    pattern = north west lines,
    fill = blue0] table [y=sgemm, meta=Label,x expr=\coordindex] {\datagsuitespmmgin};
    \addplot [draw = blue4,
    semithick,
    pattern = grid,
    fill = blue4] table [y=other, meta=Label,x expr=\coordindex] {\datagsuitespmmgin};
    \end{axis}
    

    \end{tikzpicture}
    
    \begin{tikzpicture}

    \begin{axis}[
    width=3cm,
    height=1.7cm,
    title style = {font=\footnotesize},
    axis line style={draw=none},
    ticks = none,
    ybar stacked,   
    ymin=0,         
    legend style={at={(axis cs:-7,0)},anchor=south west, font=\footnotesize, /tikz/every even column/.append style={column sep=0.2cm}},
    legend columns=5,
    ]
    \addplot [draw = blue0,
    semithick,
    pattern = north west lines,
    fill = blue0] table [y=sgemm, meta=Label,x expr=\coordindex] {\datalegend};   
    \addplot [draw = blue2,
    semithick,
    pattern = crosshatch,
    fill = blue2] table [y=scatter, meta=Label,x expr=\coordindex] {\datalegend};
    \addplot [draw = blue1,
    semithick,
    pattern = north west lines,
    pattern color = blue1] table [y=indexSelect, meta=Label,x expr=\coordindex] {\datalegend};
    \addplot [draw = blue3,
    semithick,
    pattern = crosshatch,
    pattern color=blue3] table [y=SpMM, meta=Label,x expr=\coordindex] {\datalegend};
    \addplot [draw = blue4,
    semithick,
    pattern = grid,
    fill = blue4] table [y=other, meta=Label,x expr=\coordindex] {\datalegend};
    \legend{sgemm,scatter,indexSelect,SpMM,other}
    \end{axis}

    \end{tikzpicture}

    \caption{Execution time distribution of the kernels with different GNN frameworks.}
    \label{fig:exec-time-dist-kernels}

\end{figure*}

We also show the execution time distribution of the kernels in Fig.~\ref{fig:exec-time-dist-kernels}.
gSuite shows a similar distribution to that of PyG and DGL.
We observed that the GNN model is the main determinative factor for the distribution of kernel execution times.




\subsubsection{Instruction Breakdown}
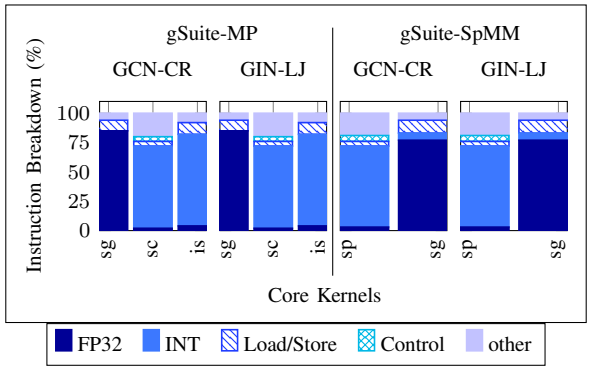
\begin{figure}[h]
  \centering
  
  \pgfplotsset{every tick label/.append style={font=\footnotesize}}

\pgfplotstableread{ 
Label           FP32    INT     Load/Store      Control     other
sgemm           85      0       9               0           6
scatter         2       70      4               4           20
indexSelect     4       78      10              0           8      
    }\instrbrgsuitemp
    
\pgfplotstableread{ 
Label           FP32    INT     Load/Store      Control     other
sgemm           81      0       11               0           8
scatter         2       68      4               4           22
indexSelect     4       78      10              0           8      
    }\instrbrgsuitempgin
    
\pgfplotstableread{ 
Label           FP32    INT     Load/Store      Control     other
sgemm           75      0       19               0           6
scatter         2       70      4               4           20
indexSelect     9       68      15              0           8
    }\instrbrgsuitempsag

\pgfplotstableread{ 
Label           FP32    INT     Load/Store      Control     other
SpMM            3       69      4               5           19
sgemm           77      6       11              0           6
    }\instrbrgsuitespmmgcn
    
\pgfplotstableread{ 
Label           FP32    INT     Load/Store      Control     other
SpMM            3       59      4               10           24
sgemm           67      6       21              0           6
    }\instrbrgsuitespmmgin
    
\pgfplotstableread{ 
Label           FP32    INT     Load/Store      Control     other
SpMM            3       69      4               5           19
sgemm           77      6       11              0           6
    }\instrbrgsuitespmmsag

 \begin{tikzpicture}[framed]


    \node at (1.5,2.6) {\footnotesize gSuite-MP};
    \node at (4.8,2.6) {\footnotesize gSuite-SpMM};
    
    \draw (3.1, 2.7) -- (3.1, -0.6);
    
    \node at (-0.85,1) {\rotatebox{90}{\footnotesize{Instruction Breakdown (\%)}}};
    \node at (3,-0.85) {\footnotesize Core Kernels};

    \begin{axis}[
    width=3cm,
    height=3.3cm,
    title={GCN-CR},
    title style = {font=\footnotesize},
    ybar stacked,   
    bar width = 0.51cm,
    ymin=0,         
    xtick=data,     
    legend style={at={(axis cs:105,0)},anchor=south west, font=\footnotesize},
    ytick={0,25,50,75,100},
    xticklabels = {sg,sc,is},  
    xticklabel style={rotate=90}
    ]
    \addplot [draw = blue0,
    semithick,
    pattern = north west lines,
    fill = blue0] table [y=FP32, meta=Label,x expr=\coordindex] {\instrbrgsuitemp};   
    \addplot [draw = blue2,
    semithick,
    pattern = crosshatch,
    fill = blue2] table [y=INT, meta=Label,x expr=\coordindex] {\instrbrgsuitemp};
    \addplot [draw = blue1,
    semithick,
    pattern = north west lines,
    pattern color=blue1] table [y=Load/Store, meta=Label,x expr=\coordindex] {\instrbrgsuitemp};
    \addplot [draw = blue3,
    semithick,
    pattern = crosshatch,
    pattern color=blue3] table [y=Control, meta=Label,x expr=\coordindex] {\instrbrgsuitemp};
    \addplot [draw = blue4,
    semithick,
    pattern = grid,
    fill = blue4] table [y=other, meta=Label,x expr=\coordindex] {\instrbrgsuitemp};
    \end{axis}

    \begin{axis}[
    width=3cm,
    height=3.3cm,
    xshift=1.6cm,
    title={GIN-LJ},
    title style = {font=\footnotesize},
    ybar stacked,   
    bar width = 0.51cm,
    ymin=0,         
    xtick=data,     
    legend style={at={(axis cs:105,0)},anchor=south west, font=\footnotesize},
    ytick={0,25,50,75,100},
    xticklabels = {sg,sc,is},  
    yticklabels={,,},
    xticklabel style={rotate=90}
    ]
    \addplot [draw = blue0,
    semithick,
    pattern = north west lines,
    fill = blue0] table [y=FP32, meta=Label,x expr=\coordindex] {\instrbrgsuitemp};   
    \addplot [draw = blue2,
    semithick,
    pattern = crosshatch,
    fill = blue2] table [y=INT, meta=Label,x expr=\coordindex] {\instrbrgsuitemp};
    \addplot [draw = blue1,
    semithick,
    pattern = north west lines,
    pattern color=blue1] table [y=Load/Store, meta=Label,x expr=\coordindex] {\instrbrgsuitemp};
    \addplot [draw = blue3,
    semithick,
    pattern = crosshatch,
    pattern color=blue3] table [y=Control, meta=Label,x expr=\coordindex] {\instrbrgsuitemp};
    \addplot [draw = blue4,
    semithick,
    pattern = grid,
    fill = blue4] table [y=other, meta=Label,x expr=\coordindex] {\instrbrgsuitemp};
    \end{axis}

    \begin{axis}[
    width=3cm,
    height=3.3cm,
    xshift=3.2cm,
    title={GCN-CR},
    title style = {font=\footnotesize},
    ybar stacked,   
    bar width = 1.05cm,
    ymin=0,         
    xtick=data,     
    legend style={at={(axis cs:105,0)},anchor=south west, font=\footnotesize},
    yticklabels={,,},
    xticklabels = {sp,sg},  
    xticklabel style={rotate=90}
    ]
    \addplot [draw = blue0,
    semithick,
    pattern = north west lines,
    fill = blue0] table [y=FP32, meta=Label,x expr=\coordindex] {\instrbrgsuitespmmgcn};   
    \addplot [draw = blue2,
    semithick,
    pattern = crosshatch,
    fill = blue2] table [y=INT, meta=Label,x expr=\coordindex] {\instrbrgsuitespmmgcn};
    \addplot [draw = blue1,
    semithick,
    pattern = north west lines,
    pattern color=blue1] table [y=Load/Store, meta=Label,x expr=\coordindex] {\instrbrgsuitespmmgcn};
    \addplot [draw = blue3,
    semithick,
    pattern = crosshatch,
    pattern color=blue3] table [y=Control, meta=Label,x expr=\coordindex] {\instrbrgsuitespmmgcn};
    \addplot [draw = blue4,
    semithick,
    pattern = grid,
    fill = blue4] table [y=other, meta=Label,x expr=\coordindex] {\instrbrgsuitespmmgcn};
    \end{axis}

    \begin{axis}[
    width=3cm,
    height=3.3cm,
    xshift=4.8cm,
    title={GIN-LJ},
    title style = {font=\footnotesize},
    ybar stacked,   
    bar width = 1.05cm,
    ymin=0,         
    xtick=data,     
    legend style={at={(axis cs:-7,0)},anchor=south west, font=\footnotesize,
    /tikz/every even column/.append style={column sep=0.2cm}},
    yticklabels={,,},
    xticklabels = {sp,sg},  
    xticklabel style={rotate=90}
    ]
    \addplot [draw = blue0,
    semithick,
    pattern = north west lines,
    fill = blue0] table [y=FP32, meta=Label,x expr=\coordindex] {\instrbrgsuitespmmgcn};   
    \addplot [draw = blue2,
    semithick,
    pattern = crosshatch,
    fill = blue2] table [y=INT, meta=Label,x expr=\coordindex] {\instrbrgsuitespmmgcn};
    \addplot [draw = blue1,
    semithick,
    pattern = north west lines,
    pattern color=blue1] table [y=Load/Store, meta=Label,x expr=\coordindex] {\instrbrgsuitespmmgcn};
    \addplot [draw = blue3,
    semithick,
    pattern = crosshatch,
    pattern color=blue3] table [y=Control, meta=Label,x expr=\coordindex] {\instrbrgsuitespmmgcn};
    \addplot [draw = blue4,
    semithick,
    pattern = grid,
    fill = blue4] table [y=other, meta=Label,x expr=\coordindex] {\instrbrgsuitespmmgcn};
    \end{axis}

    \end{tikzpicture}

    \pgfplotstableread{ 
Label           FP32    INT     Load/Store      Control     other
SpMM            0       0       0               0           0
}\datalegendinstr
    
    \begin{tikzpicture}

    \begin{axis}[
    width=3cm,
    height=1.7cm,
    title style = {font=\footnotesize},
    axis line style={draw=none},
    ticks = none,
    ybar stacked,   
    ymin=0,         
    legend style={at={(axis cs:-7,0)},anchor=south west, font=\footnotesize,
    /tikz/every even column/.append style={column sep=0.2cm}},
    legend columns=5,
    ]
    \addplot [draw = blue0,
    semithick,
    pattern = north west lines,
    fill = blue0] table [y=FP32, meta=Label,x expr=\coordindex] {\datalegendinstr};   
    \addplot [draw = blue2,
    semithick,
    pattern = crosshatch,
    fill = blue2] table [y=INT, meta=Label,x expr=\coordindex] {\datalegendinstr};
    \addplot [draw = blue1,
    semithick,
    pattern = north west lines,
    pattern color = blue1] table [y=Load/Store, meta=Label,x expr=\coordindex] {\datalegendinstr};
    \addplot [draw = blue3,
    semithick,
    pattern = crosshatch,
    pattern color=blue3] table [y=Control, meta=Label,x expr=\coordindex] {\datalegendinstr};
    \addplot [draw = blue4,
    semithick,
    pattern = grid,
    fill = blue4] table [y=other, meta=Label,x expr=\coordindex] {\datalegendinstr};
    \legend{FP32,INT,Load/Store,Control,other}
    \end{axis}

    \end{tikzpicture}
    
    \caption{Instruction breakdown of the kernels during the execution.}
    \label{fig:instr-breakdown}

\end{figure}

Each core kernel consists of different types of instructions to accomplish its task during the execution.
We have found that each core kernel has a characteristic distribution of instructions that does not vary even though when GNN model or dataset is adjusted.
Fig~\ref{fig:instr-breakdown} shows the instruction breakdown of kernels on different models and datasets, implying that the distribution is not affected by the adjustment of GNN model and dataset.

From our instruction breakdown analysis (Fig.~~\ref{fig:instr-breakdown}), we observe that scatter and indexSelect kernels are dominated with integer operations. Because, these two kernels mainly perform address calculations for data accesses. On the other hand, sgemm kernel is highly dominated by floating operations. Based on these observations, we can suggest researchers to investigate co-scheduling of kernels and also focus on warp scheduling studies for better utilization of the functional units.

\subsubsection{Issue Stall Distribution}
\begin{figure*}[h]
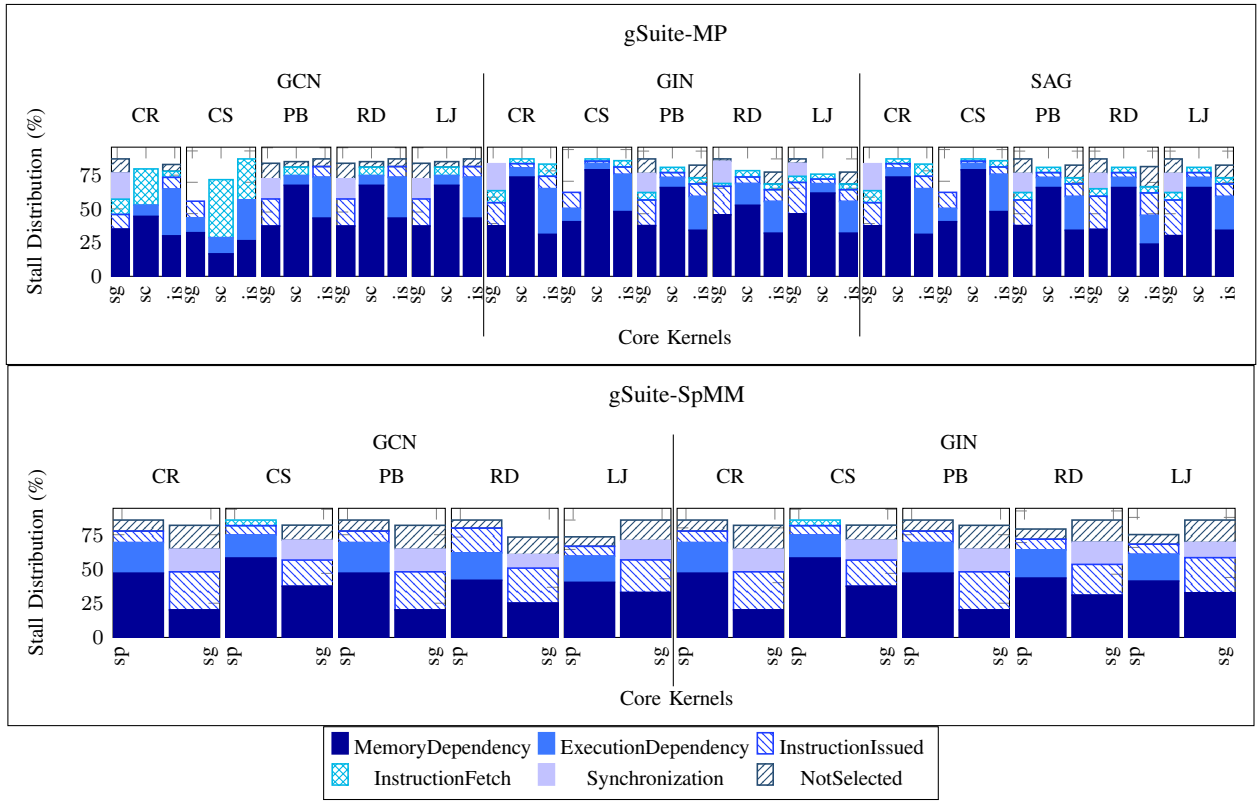

  \centering
  
  \pgfplotsset{every tick label/.append style={font=\footnotesize}}

\pgfplotstableread{ 
Label           MemoryDependency ExecutionDependency InstructionIssued InstructionFetch Synchronization NotSelected
sgemm           35  0  10.89  11.31  19.36  10.48
scatter         44.44  8.33  0  26.86  0  0
indexSelect     30  34.66  8.73  4.5  0  5.02
    }\issuestallgsuitempgcncora
    
\pgfplotstableread{ 
Label           MemoryDependency ExecutionDependency InstructionIssued InstructionFetch Synchronization NotSelected
sgemm           34.8  11.89  13.22  0  0  0
scatter         17.92  12.92  0  46.25  0  0
indexSelect     28.42  32.63  0  32.58  0  0
    }\issuestallgsuitempgcnciteseer
    
\pgfplotstableread{ 
Label           MemoryDependency ExecutionDependency InstructionIssued InstructionFetch Synchronization NotSelected
sgemm           39.11  0  21.07  0  15.65  12.1
scatter         70.77  7.62  0  6.55  0  4.26
indexSelect     45.29  31.75  8.35  0  0  5.94
    }\issuestallgsuitempgcnpubmed
    
\pgfplotstableread{ 
Label           MemoryDependency ExecutionDependency InstructionIssued InstructionFetch Synchronization NotSelected
sgemm           46.3  0  23  0  11.9  7.2
scatter         70.77  7.62  0  6.55  0  4.26
indexSelect     47.3  30.6  8.35  0  0  5.94
    }\issuestallgsuitempgcnreddit
    
\pgfplotstableread{ 
Label           MemoryDependency ExecutionDependency InstructionIssued InstructionFetch Synchronization NotSelected
sgemm           49.2  0  24.2  0  8.6  10.3
scatter         70.77  7.62  0  6.55  0  4.26
indexSelect     41.8  27.6  12.4  0  0  6.1
    }\issuestallgsuitempgcnlivejournal

\pgfplotstableread{ 
Label           MemoryDependency ExecutionDependency InstructionIssued InstructionFetch Synchronization NotSelected
sgemm           38.32  0  17.54  9.11  20.42  0
scatter         75.25  6.86  3.34  3.59  0  0
indexSelect     31.82  34.63  9.43  9.22  0  0
    }\issuestallgsuitempgincora
    
\pgfplotstableread{ 
Label           MemoryDependency ExecutionDependency InstructionIssued InstructionFetch Synchronization NotSelected
sgemm           43.07  10.43  12.74  0  0  0
scatter         84.02  4.93  1.94  1.72  0  0
indexSelect     51.07  29.35  5.84  4.97  0  0
    }\issuestallgsuitempginciteseer
    
\pgfplotstableread{ 
Label           MemoryDependency ExecutionDependency InstructionIssued InstructionFetch Synchronization NotSelected
sgemm           40.47  0  20.49  5.99  15.16  11.64
scatter         70.89  8.02  3.92  4.2  0  0
indexSelect     36.77  26.98  10.04  4.92  0  10.07
    }\issuestallgsuitempginpubmed
    
\pgfplotstableread{ 
Label           MemoryDependency ExecutionDependency InstructionIssued InstructionFetch Synchronization NotSelected
sgemm           52.4  0  24.3  2.4  18.9  2
scatter         60.5  18.5  5.6  5.4  0  0
indexSelect     36.77  26.98  10.04  4.92  0  10.07
    }\issuestallgsuitempginreddit
    
\pgfplotstableread{ 
Label           MemoryDependency ExecutionDependency InstructionIssued InstructionFetch Synchronization NotSelected
sgemm           53.2  0  26.8  5.2  11  3.8
scatter         70.89  8.02  3.92  4.2  0  0
indexSelect     36.77  26.98  10.04  4.92  0  10.07
    }\issuestallgsuitempginlivejournal

\pgfplotstableread{ 
Label           MemoryDependency ExecutionDependency InstructionIssued InstructionFetch Synchronization NotSelected
sgemm           38.32  0  17.54  9.11  20.42  0
scatter         75.25  6.86  3.34  3.59  0  0
indexSelect     31.82  34.63  9.43  9.22  0  0
    }\issuestallgsuitempsagcora
    
\pgfplotstableread{ 
Label           MemoryDependency ExecutionDependency InstructionIssued InstructionFetch Synchronization NotSelected
sgemm           43.07  10.43  12.74  0  0  0
scatter         84.02  4.93  1.94  1.72  0  0
indexSelect     51.07  29.35  5.84  4.97  0  0
    }\issuestallgsuitempsagciteseer
    
\pgfplotstableread{ 
Label           MemoryDependency ExecutionDependency InstructionIssued InstructionFetch Synchronization NotSelected
sgemm           40.47  0  20.49  5.99  15.16  11.64
scatter         70.89  8.02  3.92  4.2  0  0
indexSelect     36.77  26.98  10.04  4.92  0  10.07
    }\issuestallgsuitempsagpubmed
    
\pgfplotstableread{ 
Label           MemoryDependency ExecutionDependency InstructionIssued InstructionFetch Synchronization NotSelected
sgemm           37.47  0  26.49  5.99  12.16  11.64
scatter         70.89  8.02  3.92  4.2  0  0
indexSelect     25.7  22.9  18.04  4.92  0  16.07
    }\issuestallgsuitempsagreddit
    
\pgfplotstableread{ 
Label           MemoryDependency ExecutionDependency InstructionIssued InstructionFetch Synchronization NotSelected
sgemm           32.47  0  28.49  5.99  15.16  11.64
scatter         70.89  8.02  3.92  4.2  0  0
indexSelect     36.77  26.98  10.04  4.92  0  10.07
    }\issuestallgsuitempsaglivejournal

\pgfplotstableread{ 
Label           MemoryDependency ExecutionDependency InstructionIssued InstructionFetch Synchronization NotSelected
SpMM            46.94  22.29  8.43  0  0  8.04
sgemm           20.13  0  27.8  0  16.44  17.48
    }\issuestallgsuitespmmgcncora
    
\pgfplotstableread{ 
Label           MemoryDependency ExecutionDependency InstructionIssued InstructionFetch Synchronization NotSelected
SpMM            61.99  17.9  7.33  4.38  0  0
sgemm           39.97  0  20.41  0  15.48  11.77
    }\issuestallgsuitespmmgcnciteseer
    
\pgfplotstableread{ 
Label           MemoryDependency ExecutionDependency InstructionIssued InstructionFetch Synchronization NotSelected
SpMM            46.94  22.29  8.43  0  0  8.04
sgemm           20.13  0  27.8  0  16.44  17.48
    }\issuestallgsuitespmmgcnpubmed
    
\pgfplotstableread{ 
Label           MemoryDependency ExecutionDependency InstructionIssued InstructionFetch Synchronization NotSelected
SpMM            56.94  27.29  24.67  0  0  8.04
sgemm           34.28  0  34.8  0  13.44  17.48
    }\issuestallgsuitespmmgcnreddit
    
\pgfplotstableread{ 
Label           MemoryDependency ExecutionDependency InstructionIssued InstructionFetch Synchronization NotSelected
SpMM            46.94  22.29  8.43  0  0  8.04
sgemm           38.28  0  27.8  0  16.44  17.48
    }\issuestallgsuitespmmgcnlivejournal

\pgfplotstableread{ 
Label           MemoryDependency ExecutionDependency InstructionIssued InstructionFetch Synchronization NotSelected
SpMM            46.94  22.29  8.43  0  0  8.04
sgemm           20.13  0  27.8  0  16.44  17.48
    }\issuestallgsuitespmmgincora
    
\pgfplotstableread{ 
Label           MemoryDependency ExecutionDependency InstructionIssued InstructionFetch Synchronization NotSelected
SpMM            61.99  17.9  7.33  4.38  0  0
sgemm           39.97  0  20.41  0  15.48  11.77
    }\issuestallgsuitespmmginciteseer
    
\pgfplotstableread{ 
Label           MemoryDependency ExecutionDependency InstructionIssued InstructionFetch Synchronization NotSelected
SpMM            46.94  22.29  8.43  0  0  8.04
sgemm           20.13  0  27.8  0  16.44  17.48
    }\issuestallgsuitespmmginpubmed
    
\pgfplotstableread{ 
Label           MemoryDependency ExecutionDependency InstructionIssued InstructionFetch Synchronization NotSelected
SpMM            46.94  22.29  8.43  0  0  8.04
sgemm           33.46  0  24.3  0  17.56  17.48
    }\issuestallgsuitespmmginreddit
    
\pgfplotstableread{ 
Label           MemoryDependency ExecutionDependency InstructionIssued InstructionFetch Synchronization NotSelected
SpMM            46.94  22.29  8.43  0  0  8.04
sgemm           37.18  0  29.3  0  12.6  18.7
    }\issuestallgsuitespmmginlivejournal

\pgfplotstableread{ 
Label           MemoryDependency ExecutionDependency InstructionIssued InstructionFetch Synchronization NotSelected
SpMM            49.80  21.11  7.96  5.77  0  7.53
sgemm           28.90  2.31  23.39  10.23  17.65  11.80
    }\issuestallgsuitespmmsagcora
    
\pgfplotstableread{ 
Label           MemoryDependency ExecutionDependency InstructionIssued InstructionFetch Synchronization NotSelected
SpMM            61.99  17.9  7.33  4.38  0  0
sgemm           39.97  0  20.41  0  15.48  11.77
    }\issuestallgsuitespmmsagciteseer
    
\pgfplotstableread{ 
Label           MemoryDependency ExecutionDependency InstructionIssued InstructionFetch Synchronization NotSelected
SpMM            61.34  17.80  7.44  4.50  0  4.36
sgemm           38.28  3.10  21.14  6.19  15.54  12.30
    }\issuestallgsuitespmmsagpubmed

\pgfplotstableread{ 
Label           MemoryDependency ExecutionDependency InstructionIssued InstructionFetch Synchronization NotSelected
SpMM            61.34  17.80  7.44  4.50  0  4.36
sgemm           38.28  3.10  21.14  6.19  15.54  12.30
    }\issuestallgsuitespmmsagreddit

\pgfplotstableread{ 
Label           MemoryDependency ExecutionDependency InstructionIssued InstructionFetch Synchronization NotSelected
SpMM            61.34  17.80  7.44  4.50  0  4.36
sgemm           38.28  3.10  21.14  6.19  15.54  12.30
    }\issuestallgsuitespmmsaglivejournal

\pgfplotstableread{ 
Label           FP32    INT     Load/Store      Control     other
sgemm           85      0       9               0           6
scatter         2       70      4               4           20
indexSelect     4       78      10              0           8      
    }\instrbrgsuitempgin
    
\pgfplotstableread{ 
Label           FP32    INT     Load/Store      Control     other
sgemm           85      0       9               0           6
scatter         2       70      4               4           20
indexSelect     4       78      10              0           8
    }\instrbrgsuitempsag

\pgfplotstableread{ 
Label           FP32    INT     Load/Store      Control     other
SpMM            3       69      4               5           19
sgemm           77      6       11              0           6
    }\instrbrgsuitespmmgcn
    
\pgfplotstableread{ 
Label           FP32    INT     Load/Store      Control     other
SpMM            3       69      4               5           19
sgemm           77      6       11              0           6
    }\instrbrgsuitespmmgin
    
\pgfplotstableread{ 
Label           FP32    INT     Load/Store      Control     other
SpMM            3       69      4               5           19
sgemm           77      6       11              0           6
    }\instrbrgsuitespmmsag



    \pgfplotstableread{ 
Label           MemoryDependency    ExecutionDependency     InstructionIssued      InstructionFetch     Synchronization   NotSelected
SpMM            0       0       0               0           0    0
}\datalegendinstr
    
    %
    
    \caption{Issue stall distribution of the kernels during the execution, comparing MP and SpMM kernels across different GNN models and datasets.}
    \label{fig:issue-stall-dist}

\end{figure*}

We evaluate and analyze the issue stall distribution of core GNN kernels.
Issue stalls explain why an active warp is not eligible during its execution. 
Prior studies showed that the change in the characteristics of input workload has a strong effect on GNN computation\cite{gcn-char, gnn-char}.
We observe a similar behaviour in our experiments.
As the size of the dataset gets larger, all the core kernels except sgemm develop memory dependency.
Fig.~\ref{fig:issue-stall-dist} illustrates the issue stall cycle distribution of core kernels in MP and SpMM based implementations of GCN, GIN SAG models, running with our five datasets.

We found that memory dependency is the dominant stall in both MP- and SpMM-based implementations, with 46.3\% on average. This is due to irregular memory access pattern of GNN Inference tasks.
%

\subsubsection{Warp Occupancy Distribution}
This metric stands for the ratio of active warps to maximum number of supported active warps, from GPGPU-Sim.
We use this metric to measure the utilization of functional units.
In this analysis, \emph{stall} state shows that pipeline is stalled and therefore cannot issue any instructions.
\emph{Idle} state means the warps were issued but not ready to execute next instruction.
Finally, W$X$ refers to $X$ active threads were scheduled into pipeline.

\begin{figure*}[h]
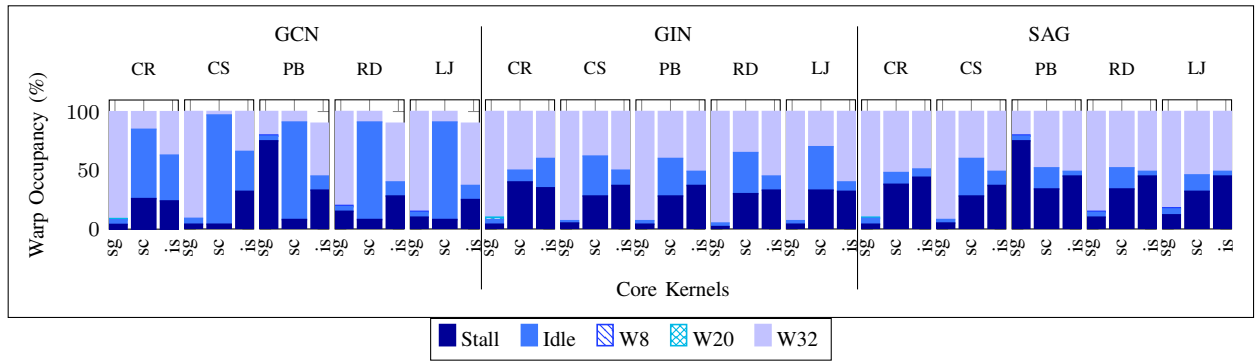

  \centering
  
  \pgfplotsset{every tick label/.append style={font=\footnotesize}}

\pgfplotstableread{ 
Label           Stall Idle W8 W20 W32
sgemm           4  4  0 1  91
scatter         26  59  0  0  15
indexSelect     24  39  0  0  37
    }\warpoccgsuitempgcncora
    
\pgfplotstableread{ 
Label           Stall Idle W8 W20 W32
sgemm           4  5  0  0  91
scatter         4  93  0  0  3
indexSelect     32  34  0  0  34
}\warpoccgsuitempgcnciteseer
    
\pgfplotstableread{ 
Label           Stall Idle W8 W20 W32
sgemm           75  4  1  0  20
scatter         8  83  0  0  9
indexSelect     33  12  0  0  45
    }\warpoccgsuitempgcnpubmed
    
\pgfplotstableread{ 
Label           Stall Idle W8 W20 W32
sgemm           15  4  1  0  80
scatter         8  83  0  0  9
indexSelect     28  12  0  0  50
    }\warpoccgsuitempgcnreddit
    
\pgfplotstableread{ 
Label           Stall Idle W8 W20 W32
sgemm           10  4  1  0  85
scatter         8  83  0  0  9
indexSelect     25  12  0  0  53
    }\warpoccgsuitempgcnlivejournal

\pgfplotstableread{ 
Label           Stall Idle W8 W20 W32
sgemm           4  4  0  2  90
scatter         40  10  0  0  50
indexSelect     35  25  0  0  40
    }\warpoccgsuitempgincora
    
\pgfplotstableread{ 
Label           Stall Idle W8 W20 W32
sgemm           5  2  0  0  93
scatter         28  34  0  0  38
indexSelect     37  13  0  0  50
    }\warpoccgsuitempginciteseer
    
\pgfplotstableread{ 
Label           Stall Idle W8 W20 W32
sgemm           4  3  0  0  93
scatter         28  32  0  0  40
indexSelect     37  12  0  0  51
    }\warpoccgsuitempginpubmed
    
\pgfplotstableread{ 
Label           Stall Idle W8 W20 W32
sgemm           2  3  0  0  95
scatter         30  35  0  0  35
indexSelect     33  12  0  0  55
    }\warpoccgsuitempginreddit
    
\pgfplotstableread{ 
Label           Stall Idle W8 W20 W32
sgemm           4  3  0  0  93
scatter         33  37  0  0  30
indexSelect     32  8  0  0  60
    }\warpoccgsuitempginlivejournal

\pgfplotstableread{ 
Label           Stall Idle W8 W20 W32
sgemm           4  5  0  1	90
scatter         38  10  0  0  52
indexSelect     44	 7	 0  0  49
    }\warpoccgsuitempsagcora
    
\pgfplotstableread{ 
Label           Stall Idle W8 W20 W32
sgemm           5  3  0  0  92
scatter         28  32  0  0  40
indexSelect     37  12  0  0  51
    }\warpoccgsuitempsagciteseer
    
\pgfplotstableread{ 
Label           Stall Idle W8 W20 W32
sgemm           75  4  1	 0  20
scatter         34  18  0  0  48
indexSelect     45  4  0  0  51
    }\warpoccgsuitempsagpubmed
    
\pgfplotstableread{ 
Label           Stall Idle W8 W20 W32
sgemm           10  4  1	 0  85
scatter         34  18  0  0  48
indexSelect     45  4  0  0  51
    }\warpoccgsuitempsagreddit
    
\pgfplotstableread{ 
Label           Stall Idle W8 W20 W32
sgemm           12  5  1	 0  82
scatter         32  14  0  0  54
indexSelect     45  4  0  0  51
    }\warpoccgsuitempsaglivejournal



    \pgfplotstableread{ 
Label           Stall Idle W8 W20 W32
SpMM            0 0 0 0 0
}\datalegendwarpocc
    
    %
    
    \caption{Warp occupancy distribution of the gSuite-MP kernels on varying GNN models and datasets.}
    \label{fig:warp-occupancy-dist}

\end{figure*}

During the experiments, we observed that the type of GNN model plays a crucial role in pipeline utilization.
MP-based kernels (scatter and indexSelect) of GCN tend to stay idle during the execution, unlike GIN and SAG kernels.
However, sgemm kernel is immune to these GNN model adjustments.
Fig.~\ref{fig:warp-occupancy-dist} shows how the utilization levels change across GNN models and our datasets.

Figures \ref{fig:issue-stall-dist}  and \ref{fig:warp-occupancy-dist} also highlights the inefficiency of front-end when running indexSelect and scatter kernels in GCN MP model (we observe high instruction fetch in Fig.~\ref{fig:issue-stall-dist}  and high idle time in Fig.~\ref{fig:warp-occupancy-dist} for these kernels in GCN MP model, especially with small sized datasets (CR and CS)).

\subsubsection{L1/L2 Cache Hit Rate}
As GNN operations draw an irregular access pattern on memory; we expect high miss rates consistent with prior characterization studies\cite{gnnmark, Subramaniyan2021}.
We aim to show how input workload characteristics affect the cache miss rates during GNN computation.

\begin{figure}[h]
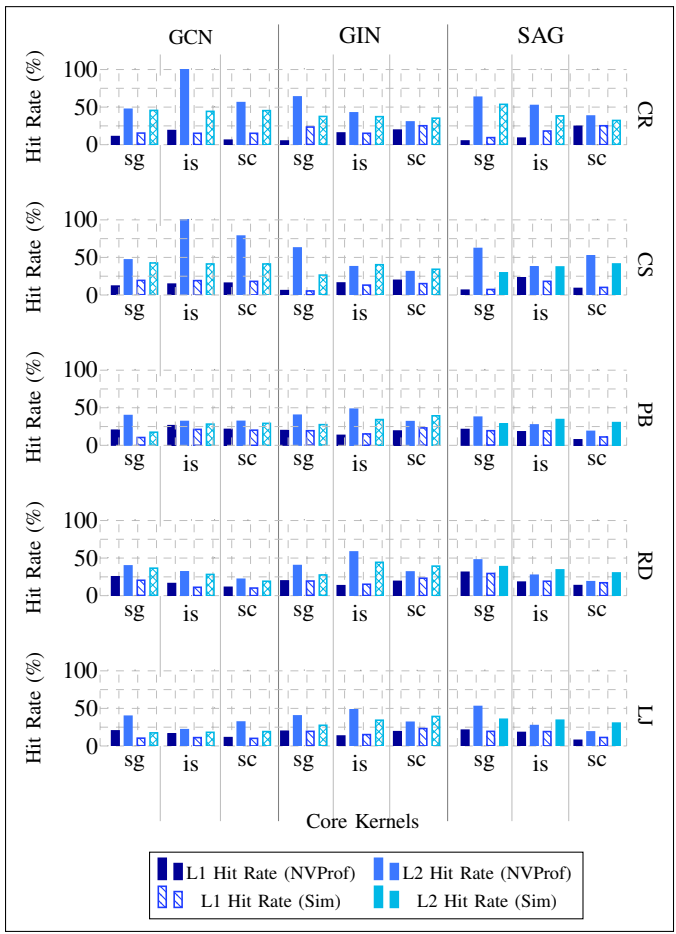

  \centering



\caption{L1 and L2 cache hit rates of the MP-gSuite kernels, comparing NVIDIA Profiler and GPGPU-Sim outcomes.}
\label{fig:l1-l2-hit-rate}
\end{figure}

Moreover, we point out resemblance and differences of a hardware profiler statistics and architectural simulator results for this metric.
We collected the cache utilization statistics by using both nvprof and GPGPU-Sim.
Fig.~\ref{fig:l1-l2-hit-rate} depicts the results of experiments and compares nvprof results with GPGPU-Sim outcomes.

We observe that L1 cache hit ratio values for profiler and simulator are more aligned than L2 cache hit values. Specifically, for some workloads (CR and CS), the simulator-indicated memory performance is not well matching with the hardware-based memory performance. This shows us that more validation study is required on GPGPU-Sim's memory model.

From our detailed analysis on L1/L2 cache accesses/hits, we observe that the GCN workloads have some or limited locality. This suggests that architects should study GNN friendly caching and prefetching options. Specifically, indexSelect kernel cannot utilize memory efficiently. Average memory utilization of 34.6\% combined with the high L1D cache miss rates we observed, we suggest researchers to investigate other caching techniques to be applied particularly on indexSelect kernel.

We also notice that larger input workloads result in less L1/L2 cache hit ratios. These extremely low L1D cache hit rates points out that caching may not be a good technique for GNN-Inference. Therefore using L1 cache bypassing techniques can be considered as an alternative to alleviate such a problem.

From issue stall distribution (Fig.~\ref{fig:issue-stall-dist}) and L1/L2 cache hit rates (Fig.~\ref{fig:l1-l2-hit-rate}), we observe that indexSelect and scatter kernels suffer from memory dependency. This suggests that considering the implementations of functionalities of these kernels on memory side would be good option in terms of energy consumption, utilization and performance. The atomic reduce operation in scatter kernel affects the performance of this kernel. Therefore, this kernel could benefit from the architectural support for more efficient synchronization operations. Architects may also investigate the prefetching options.

\subsubsection{Compute/Memory Utilization}
\begin{figure}[h]
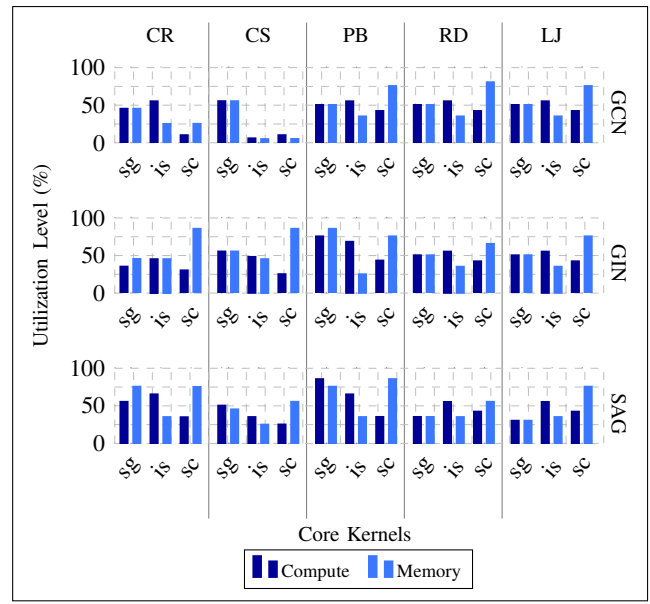

  \centering



\caption{Compute and memory utilization levels of MP-gSuite kernels on varying GNN models and datasets.}
\label{fig:compute-memory-util}
\end{figure}

We examine the performance limiter for each core kernel during the execution.
Low compute and memory utilization values point that a kernel's performance is bounded by instruction and memory latency.
We observe that scatter kernel utilizes memory more efficiently than other kernels, especially when employed in GIN and SAG.
Compute and memory utilization of sgemm kernel scales up as the input workload is bigger (e.g. with LiveJournal dataset).
These utilization levels are presented in Fig.~\ref{fig:compute-memory-util}.

\section{Related Work}
The continuous growth of real-world graph data has led to the development of new methods to process such data effectively\cite{Zhou:Mermaid:2017, Gonzalez:PowerGraph:2012, Low:GraphLab:2010, Malewicz:Pregel:2010}.
With their ability to handle high memory access bandwidth and massive parallelism, GPUs gained the attention of researchers and engineers, especially for graph processing tasks\cite{He2010, Merrill:2012:BFS, Ashari2014}.
While GPUs offer significant performance improvements for graph applications, they come with several challenges to deal with.
There are many studies on efficient implementation and performance evaluation of graph applications on GPUs\cite{Harish:2007:ALG,Luo:2010:BFS, Merrill:2012:BFS,Vineet:2009:MST, Kalentev:2011:CCL,Yonehara:2015:CCL,Auer:2012:GM, Grosset:2011:EGC}.
These studies mainly focus on data layout optimizations, memory access patterns optimizations, and workload mapping for load balancing.
There are also several graph frameworks, benchmarks and characterization studies on graph algorithms running on GPUs \cite{Hong:MultiGraph:2017, Wang:Gunrock:2016, Khorasani:CuSha:2014, Che:pannotia:2013, Xu:2014:Graph}. 

Most DNN applications and frameworks also utilize the GPUs' computing capability.
We have seen many DNN frameworks\cite{Caffe,Theano} 
and studies that measure the performance of DNN applications on GPUs\cite{Chen2017, Hadidi2019, Huang2020, Horikawa2019, dnn-train,dawn-bench,dnn-mark,rodinia}. 

Increasing interest in GNNs has led the model benchmarking \cite{Errica2019,Dwivedi2020,Mernyei2020,Fung2021} and architectural performance analysis studies for GNNs\cite{gnnmark, gnn-char,gcn-char}.
There are also efforts towards providing datasets for benchmarking GNNs \cite{Mernyei2020}.

The most relevant work to ours are \cite{MatthiasFey2019,Wang2019,gcn-char,gnn-char,hygcn,grip,gnnmark}. 

PyTorch Geometric (PyG) is a GNN framework based on PyTorch, which provides an infrastructure to build GNN pipelines with implemented GNN models and datasets.
GNN models in PyG are inherited from a base class called \emph{MessagePassing}.
Another common GNN framework is Deep Graph Library (DGL), which gives user a choice to alternate among three DNN libraries: PyTorch, Tensorflow, and MXNet.
DGL follows the \emph{SpMM} schema in its GNN implementations.

Yan et al.\cite{gcn-char} characterize GCNs at Inference level with varying workloads, using PyG.
Zhang et al.\cite{gnn-char} characterize GNN Inference on GPUs by taking two popular frameworks into consideration: PyG and DGL. They consider the most common GNN models in a stage level analysis manner, and make implications for hardware accelerators.
However, this work is not open-source and cannot be extended by research community.

GNNMark\cite{gnnmark} is a benchmark suite that is designed to understand system-level and architectural implications of GNNs, specifically during the training phase.
A range number of GNN models are covered, and many datasets are used.
They examine the scalability of GNN training across a multi-GPU system.
However, unlike our study, GNNMark is not intended to be configurable.
Workloads are tend to be treated as applications of model-dataset couples.
GNNMark is also using PyG and DGL to build GNN pipelines.

\section{Conclusion and Future Work}
In this paper, we present gSuite, a flexible and framework-independent benchmark suite for GNNs.
By providing this suite, we aim to fill the absence of a GNN-oriented benchmark utility which is not dependent to any other framework such as PyG and DGL.
As a proof of concept, we characterize and profile the computation of GNN Inference by using our proposed benchmark suite.
We utilize both a hardware profiler and a cycle accurate simulator to measure the performance of GNN computation.

We provide gSuite as an open-source project, hence all the experiments are reproducible with proper configurations\cite{gSuite}.
We also welcome any contribution and suggestion to the benchmark suite.

As a future work, we plan to extend our benchmark suite by adding support for GNN-Training, which includes the implementation of training-related aspects such as neuron layers, propagations, weights, etc. 

We also plan to support different architectures such as FPGAs and AMD GPUs by implementing our core kernels with OpenCL.


\begin{thebibliography}{00}


\bibitem{Zhou:Mermaid:2017} J. Zhou, C. Xu, X. Chen, C. Wang and X. Zhou, ``Mermaid: Integrating Vertex-Centric with Edge-Centric for Real-World Graph Processing,'' 2017 17th IEEE/ACM International Symposium on Cluster, Cloud and Grid Computing (CCGRID), 2017, pp. 780-783, doi: 10.1109/CCGRID.2017.63.

\bibitem{Gonzalez:PowerGraph:2012} Joseph E. Gonzalez, Yucheng Low, Haijie Gu, Danny Bickson, and Carlos Guestrin, ``PowerGraph: distributed graph-parallel computation on natural graphs,'' In Proceedings of the 10th USENIX conference on Operating Systems Design and Implementation (OSDI'12). USENIX Association, USA, 17–30.

\bibitem{Low:GraphLab:2010} Yucheng Low, Joseph Gonzalez, Aapo Kyrola, Danny Bickson, Carlos Guestrin, and Joseph Hellerstein, ``GraphLab: a new framework for parallel machine learning,'' In Proceedings of the Twenty-Sixth Conference on Uncertainty in Artificial Intelligence (UAI'10). AUAI Press, Arlington, Virginia, USA, 340–349.

\bibitem{Malewicz:Pregel:2010} Grzegorz Malewicz, Matthew H. Austern, Aart J.C Bik, James C. Dehnert, Ilan Horn, Naty Leiser, and Grzegorz Czajkowski, ``Pregel: a system for large-scale graph processing,'' In Proceedings of the 2010 ACM SIGMOD International Conference on Management of data (SIGMOD '10). Association for Computing Machinery, New York, NY, USA, 135–146. https://doi.org/10.1145/1807167.1807184

\bibitem{Hong:MultiGraph:2017} C. Hong, A. Sukumaran-Rajam, J. Kim and P. Sadayappan, ``MultiGraph: Efficient Graph Processing on GPUs,'' 2017 26th International Conference on Parallel Architectures and Compilation Techniques (PACT), 2017, pp. 27-40, doi: 10.1109/PACT.2017.48.

\bibitem{Wang:Gunrock:2016} Yangzihao Wang, Andrew Davidson, Yuechao Pan, Yuduo Wu, Andy Riffel, and John D. Owens,``Gunrock: a high-performance graph processing library on the GPU,'' SIGPLAN Not. 51, 8, Article 11 (August 2016), 12 pages. https://doi.org/10.1145/3016078.2851145

\bibitem{Khorasani:CuSha:2014} Farzad Khorasani, Keval Vora, Rajiv Gupta, and Laxmi N. Bhuyan, ``CuSha: vertex-centric graph processing on GPUs,'' In Proceedings of the 23rd international symposium on High-performance parallel and distributed computing (HPDC '14). Association for Computing Machinery, New York, NY, USA, 239–252. https://doi.org/10.1145/2600212.2600227

\bibitem{Che:pannotia:2013} S. Che, B. M. Beckmann, S. K. Reinhardt and K. Skadron, ``Pannotia: Understanding irregular GPGPU graph applications,'' 2013 IEEE International Symposium on Workload Characterization (IISWC), 2013, pp. 185-195, doi: 10.1109/IISWC.2013.6704684.

\bibitem{Xu:2014:Graph} Q. Xu, H. Jeon and M. Annavaram, ``Graph processing on GPUs: Where are the bottlenecks?,'' 2014 IEEE International Symposium on Workload Characterization (IISWC), 2014, pp. 140-149, doi: 10.1109/IISWC.2014.6983053.

\bibitem{Vineet:2009:MST} Vibhav Vineet, Pawan Harish, Suryakant Patidar, and P. J. Narayanan, ``Fast minimum spanning tree for large graphs on the GPU,'' In Proceedings of the Conference on High Performance Graphics 2009 (HPG '09). Association for Computing Machinery, New York, NY, USA, 167–171. https://doi.org/10.1145/1572769.1572796

\bibitem{Kalentev:2011:CCL} Oleksandr Kalentev, Abha Rai, Stefan Kemnitz, Ralf Schneider, ``Connected component labeling on a 2D grid using CUDA,'' Journal of Parallel and Distributed Computing, Volume 71, Issue 4, 2011, Pages 615-620, ISSN 0743-7315, https://doi.org/10.1016/j.jpdc.2010.10.012.

\bibitem{Yonehara:2015:CCL} K. Yonehara and K. Aizawa, ``A Line-Based Connected Component Labeling Algorithm Using GPUs,'' 2015 Third International Symposium on Computing and Networking (CANDAR), 2015, pp. 341-345, doi: 10.1109/CANDAR.2015.78.

\bibitem{Auer:2012:GM} Bas O. Fagginger Auer and Rob H. Bisseling, ``A GPU algorithm for greedy graph matching,'' Facing the Multicore-Challenge II: aspects of new paradigms and technologies in parallel computing. Springer-Verlag, Berlin, Heidelberg, 108–119.

\bibitem{Grosset:2011:EGC} Andre Vincent Pascal Grosset, Peihong Zhu, Shusen Liu, Suresh Venkatasubramanian, and Mary Hall, ``Evaluating graph coloring on GPUs,'' SIGPLAN Not. 46, 8 (August 2011), 297–298. https://doi.org/10.1145/2038037.1941597

\bibitem{Merrill:2012:BFS} Duane Merrill, Michael Garland, and Andrew Grimshaw, ``Scalable GPU graph traversal,'' SIGPLAN Not. 47, 8 (August 2012), 117–128. https://doi.org/10.1145/2370036.2145832

\bibitem{Ashari2014} A. Ashari, N. Sedaghati, J. Eisenlohr, S. Parthasarath, and P. Sadayappan, ``Fast sparse matrix-vector multiplication on GPUs for graph applications,'' in SC14: International Conference for High Performance Computing, Networking, Storage and Analysis, 2014.

\bibitem{He2010} G. He, H. Feng, C. Li, and H. Chen, ``Parallel SimRank computation on large graphs with iterative aggregation,'' in Proceedings of the 16th ACM SIGKDD international conference on Knowledge discovery and data mining - KDD ’10, 2010.

\bibitem{Luo:2010:BFS} L. Luo, M. Wong and W. Hwu, ``An effective GPU implementation of breadth-first search,'' Design Automation Conference, 2010, pp. 52-55.

\bibitem{Harish:2007:ALG} Pawan Harish and P. J. Narayanan, ``Accelerating large graph algorithms on the GPU using CUDA,'' In Proceedings of the 14th international conference on High performance computing (HiPC'07). Springer-Verlag, Berlin, Heidelberg, 197–208.

\bibitem{Chen2022} Y. Chen, Y. Hu, K. Li, C. Yeo and K. Li, ``Approximate personalized propagation for unsupervised embedding in heterogeneous graphs'', Information Sciences, vol. 600, pp. 287-300, 2022. Available: 10.1016/j.ins.2022.04.002 [Accessed 9 April 2022].

\bibitem{Zhang2022} T. Zhang, H. Shan and M. Little, ``Causal GraphSAGE: A robust graph method for classification based on causal sampling'', Pattern Recognition, vol. 128, p. 108696, 2022. Available: 10.1016/j.patcog.2022.108696 [Accessed 9 April 2022].

\bibitem{semi-supervised-gcn} Kipf, T. N. and Welling, M., ``Semi-Supervised Classification with Graph Convolutional Networks'', 2016.

\bibitem{Marius2009} M.-C. Popescu, V. E. Balas, L. Perescu-Popescu, and N. Mastorakis, ``Multilayer perceptron and neural networks,'' WSEAS Trans. Circuits and Syst., vol. 8, no. 7, pp. 579–588, 2009.

\bibitem{Peter2018} P. Battaglia et al., "Relational inductive biases, deep learning, and graph networks", arXiv.org, 2022. [Online]. Available: https://arxiv.org/abs/1806.01261. [Accessed: 09- Apr- 2022].

\bibitem{MatthiasFey2019} M. Fey and J. E. Lenssen, “Fast Graph Representation Learning with PyTorch Geometric,” in ICLR 2019 Workshop on Representation Learning on Graphs and Manifolds, New Orleans, USA, 2019.

\bibitem{Wang2019} M. Wang et al., “Deep Graph Library: A graph-centric, highly-performant package for graph neural networks,” arXiv [cs.LG], 2019.

\bibitem{Gilmer2017} J. Gilmer, S. S. Schoenholz, P. F. Riley, O. Vinyals, and G. E. Dahl, “Neural Message Passing for Quantum Chemistry,” arXiv [cs.LG], 2017.

\bibitem{Liao2018} R. Liao, M. Brockschmidt, D. Tarlow, A. L. Gaunt, R. Urtasun, and R. Zemel, “Graph partition neural networks for semi-supervised classification,” arXiv [cs.LG], 2018.

\bibitem{Li2020} X. Li et al., “Pooling Regularized graph neural network for fMRI biomarker analysis,” Med. Image Comput. Comput. Assist. Interv., vol. 12267, pp. 625–635, 2020.

\bibitem{Nettleton2013} D. F. Nettleton, “Data mining of social networks represented as graphs,” Computer Science Review, 12-Feb-2013. [Online]. Available: https://www.sciencedirect.com/science/article/pii/S1574013712000445. [Accessed: 17-May-2022].

\bibitem{hygcn} M. Yan et al., "HyGCN: A GCN Accelerator with Hybrid Architecture," 2020 IEEE International Symposium on High Performance Computer Architecture (HPCA), 2020, pp. 15-29, doi: 10.1109/HPCA47549.2020.00012.

\bibitem{grip} Kevin Kiningham, Christopher Re and Philip Levis, ``GRIP: A Graph Neural Network Accelerator Architecture,'' arXiv, 2020. doi: 10.48550/ARXIV.2007.13828.

\bibitem{gcn-char} Mingyu Yan, Zhaodong Chen, Lei Deng, Xiaochun Ye, Zhimin Zhang, Dongrui Fan and Yuan Xie, ``Characterizing and Understanding GCNs on GPU,'' in IEEE Computer Architecture Letters, vol. 19, no. 1, pp. 22-25, 1 Jan.-June 2020, doi: 10.1109/LCA.2020.2970395.

\bibitem{gnn-char} Zhihui Zhang, Jingwen Leng, Lingxiao Ma, Youshan Miao, Chao Li and Minyi Guo, ``Architectural Implications of Graph Neural Networks,'' in IEEE Computer Architecture Letters, vol. 19, no. 1, pp. 59-62, 1 Jan.-June 2020, doi: 10.1109/LCA.2020.2988991.

\bibitem{gnnmark} Trinayan Baruah, Kaustubh Shivdikar, Shi Dong, Yifan Sun, Saiful A Mojumder, Kihoon Jung, José L. Abellán, Yash Ukidave, Ajay Joshi, John Kim and David Kaeli, ``GNNMark: A Benchmark Suite to Characterize Graph Neural Network Training on GPUs,'' 2021 IEEE International Symposium on Performance Analysis of Systems and Software (ISPASS), 2021, pp. 13-23, doi: 10.1109/ISPASS51385.2021.00013.

\bibitem{Subramaniyan2021} A. Subramaniyan et al., "GenomicsBench: A Benchmark Suite for Genomics," 2021 IEEE International Symposium on Performance Analysis of Systems and Software (ISPASS), 2021, pp. 1-12, doi: 10.1109/ISPASS51385.2021.00012.

\bibitem{gpgpu-sim} Jonathan Lew, Deval A. Shah, Suchita Pati, Shaylin Cattell, Mengchi Zhang, Amruth Sandhupatla, Christopher Ng, Negar Goli, Matthew D. Sinclair, Timothy G. Rogers, and Tor M. Aamodt, ``Analyzing Machine Learning Workloads Using a Detailed GPU Simulator,'' 2019 IEEE International Symposium on Performance Analysis of Systems and Software (ISPASS), 2019, pp. 151-152, doi: 10.1109/ISPASS.2019.00028.

\bibitem{Scarselli2009} F. Scarselli, M. Gori, A. C. Tsoi, M. Hagenbuchner, and G. Monfardini, “The graph neural network model,” IEEE Trans. Neural Netw., vol. 20, no. 1, pp. 61–80, 2009.

\bibitem{Shi2018} 
Xuanhua Shi, Zhigao Zheng, Yongluan Zhou, Hai Jin, Ligang He, Bo Liu, and Qiang-Sheng Hua, “Graph processing on GPUs: A survey,” ACM Comput. Surv., vol. 50, no. 6, pp. 1–35, 2018.

\bibitem{Hamilton2017} W. L. Hamilton, R. Ying, and J. Leskovec, “Inductive representation learning on large graphs,” arXiv [cs.SI], 2017.

\bibitem{Weisfeiler1968} Weisfeiler, Boris, and Andrei Leman, ``The reduction of a graph to canonical form and the algebra which appears therein,'' NTI, Series 2.9 (1968): 12-16.

\bibitem{Xu2019} K. Xu, W. Hu, J. Leskovec, and S. Jegelka, “How Powerful are Graph Neural Networks?,” arXiv [cs.LG], 2018.

\bibitem{LJdataset1} L. Backstrom, D. Huttenlocher, J. Kleinberg, X. Lan. Group Formation in Large Social Networks: Membership, Growth, and Evolution. KDD, 2006.

\bibitem{LJdataset2} J. Leskovec, K. Lang, A. Dasgupta, M. Mahoney. Community Structure in Large Networks: Natural Cluster Sizes and the Absence of Large Well-Defined Clusters. Internet Mathematics 6(1) 29--123, 2009.

\bibitem{Chen2017} Yu-Hsin Chen, Tushar Krishna, Joel S. Emer and Vivienne Sze, ``Eyeriss: An Energy-Efficient Reconfigurable Accelerator for Deep Convolutional Neural Networks,'' in IEEE Journal of Solid-State Circuits, vol. 52, no. 1, pp. 127-138, Jan. 2017, doi: 10.1109/JSSC.2016.2616357.

\bibitem{Huang2020} Shanjiaoyang Huang, Weiqi Peng, Zhiwei Jia and Zhuowen Tu, ``One-Pixel Signature: Characterizing CNN Models for Backdoor Detection,'' Computer Vision – ECCV 2020. Springer International Publishing, pp. 326–341, 2020. doi: 10.1007/978-3-030-58583-9\_20.

\bibitem{dnn-train} Hongyu Zhu, Mohamed Akrout, Bojian Zheng, Andrew Pelegris, Anand Jayarajan, Amar Phanishayee, Bianca Schroeder, Gennady Pekhimenko, “Benchmarking and analyzing deep neural network training,” in 2018 IEEE International Symposium on Workload Characterization (IISWC), 2018.

\bibitem{dawn-bench} Cody A. Coleman, Deepak Narayanan, Daniel Kang, Tian Zhao, Jian Zhang, Luigi Nardi, Peter Bailis, Kunle Olukotun, Christopher R{\'e} and Matei A. Zaharia, ``DAWNBench : An End-to-End Deep Learning Benchmark and Competition,'' 2017. [Online]. Available: https://cs.stanford.edu/~matei/papers/2018/sysml\_dawnbench.pdf. [Accessed: 18-May-2022].

\bibitem{Horikawa2019} Tomoyasu Horikawa, Shuntaro C. Aoki, Mitsuaki Tsukamoto and Yukiyasu Kamitani, ``Characterization of deep neural network features by decodability from human brain activity,'' Scientific Data, vol. 6, no. 1. Springer Science and Business Media LLC, Feb. 12, 2019. doi: 10.1038/sdata.2019.12.

\bibitem{Hadidi2019} Ramyad Hadidi, Jiashen Cao, Yilun Xie, Bahar Asgari, Tushar Krishna and Hyesoon Kim, ``Characterizing the Deployment of Deep Neural Networks on Commercial Edge Devices,'' 2019 IEEE International Symposium on Workload Characterization (IISWC), 2019, pp. 35-48, doi: 10.1109/IISWC47752.2019.9041955.

\bibitem{dnn-mark} Shi Dong and David Kaeli, ``DNNMark: A Deep Neural Network Benchmark Suite for GPUs,'' In Proceedings of the General Purpose GPUs (GPGPU-10). Association for Computing Machinery, New York, NY, USA, 63–72, 2020. doi: https://doi.org/10.1145/3038228.3038239

\bibitem{rodinia} Shuai Che, Michael Boyer, Jiayuan Meng, David Tarjan, Jeremy W. Sheaffer, Sang- Ha Lee, and Kevin Skadron. 2009. Rodinia: A Benchmark Suite for Heterogeneous Computing. Proceedings of the 2009 IEEE International Symposium on Workload Characterization (IISWC) (2009), 44–54.

\bibitem{Errica2019} Federico Errica, Marco Podda, Davide Bacciu and Alessio Micheli, ``A Fair Comparison of Graph Neural Networks for Graph Classification,’' 2020 International Conference on Learning Representations (ICLR), 2020.

\bibitem{Dwivedi2020} Vijay Prakash Dwivedi, Chaitanya K. Joshi, Anh Tuan Luu, Thomas Laurent, Yoshua Bengio and Xavier Bresson, ``Benchmarking Graph Neural Networks.'' arXiv, 2020. doi: 10.48550/ARXIV.2003.00982.

\bibitem{Mernyei2020} Péter Mernyei and Cătălina Cangea, ``Wiki-CS: A Wikipedia-Based Benchmark for Graph Neural Networks,'' 2020 Graph Representation Learning and Beyond workshop (ICML 2020). arXiv, 2020. doi: 10.48550/ARXIV.2007.02901.

\bibitem{Fung2021} Victor Fung, Jiaxin Zhang, Eric Juarez and Bobby G. Sumpter, ``Benchmarking Graph Neural Networks for Materials Chemistry.'' American Chemical Society (ACS), Jan. 22, 2021. doi: 10.26434/chemrxiv.13615421.v2.

\bibitem{Rusek2019} Krzysztof Rusek and Piotr Cholda, ``Message-Passing Neural Networks Learn Little’s Law,'' IEEE Communications Letters, vol. 23, no. 2. Institute of Electrical and Electronics Engineers (IEEE), pp. 274–277, Feb. 2019. doi: 10.1109/lcomm.2018.2886259.

\bibitem{Fan2019} Wenqi Fan, Yao Ma, Qing Li, Yuan He, Eric Zhao, Jiliang Tang and Dawei Yin, ``Graph Neural Networks for Social Recommendation.'' arXiv, 2019. doi: 10.48550/ARXIV.1902.07243.

\bibitem{Chen2018} Jie Chen, Tengfei Ma and Cao Xiao, ``FastGCN: Fast Learning with Graph Convolutional Networks via Importance Sampling,'' 6th International Conference on Learning Representations, ICLR 2018, Vancouver, BC, Canada, April 30 - May 3, 2018. doi: 10.48550/ARXIV.1801.10247.

\bibitem{Velickovic2018} Petar Veličković, Guillem Cucurull, Arantxa Casanova, Adriana Romero, Pietro Liò and Yoshua Bengio, ``Graph Attention Networks,'' 6th International Conference on Learning Representations, ICLR 2018, Vancouver, BC, Canada, April 30 - May 3, 2018. doi: 10.48550/ARXIV.1710.10903.

\bibitem{Li2018} Ruoyu Li, Sheng Wang, Feiyun Zhu and Junzhou Huang, ``Adaptive Graph Convolutional Neural Networks,'' 2018 The Thirty-Second AAAI Conference on Artificial Intelligence (AAAI 18). doi:  10.48550/arXiv.1710.10903.

\bibitem{Zhang_Cui_Neumann_Chen_2018} Muhan Zhang, Zhicheng  Cui, Marion Neumann, and Yixin Chen, ``An End-to-End Deep Learning Architecture for Graph Classification'', AAAI, vol. 32, no. 1, Apr. 2018.

\bibitem{Bandyopadhyay2021}  
Sambaran Bandyopadhyay, Manasvi Aggarwal and M. Narasimha Murty, ``A Deep Hybrid Pooling Architecture for Graph Classification with Hierarchical Attention,'' Advances in Knowledge Discovery and Data Mining. Springer International Publishing, pp. 554–565, 2021. doi: 10.1007/978-3-030-75762-5\_44.

\bibitem{Yao2022} Shuanglong Yao, Dechang Pi and Junfu Chen, ``Knowledge embedding via hyperbolic skipped graph convolutional networks,'' Neurocomputing, vol. 480. Elsevier BV, pp. 119–130, Apr. 2022. doi: 10.1016/j.neucom.2022.01.037.

\bibitem{Sun2022} Dengdi Sun, Kang Yang and Zhuanlian Ding, ``Confidence-Based Simple Graph Convolutional Networks for Face Clustering,'' IEEE Access, vol. 10. Institute of Electrical and Electronics Engineers (IEEE), pp. 6459–6469, 2022. doi: 10.1109/access.2022.3142922.

\bibitem{cora} Andrew Kachites McCallum, Kamal Nigam, Jason Rennie and Kristie Seymore, ``Automating the Construction of Internet Portals with Machine Learning,'' Information Retrieval, vol. 3, no. 2. Springer Science and Business Media LLC, pp. 127–163, 2000. doi: 10.1023/a:1009953814988.

\bibitem{reddit} William L. Hamilton*, Justine Zhang*, Cristian Danescu-Niculescu-Mizil, Dan Jurafsky, Jure Leskovec. ``Loyalty in Online Communities,'' (Currently under review at WWW 2017). *Equal contribution.

\bibitem{pubmed} Ryan A. Rossi and Nesreen K Ahmed, ``The network data repository with interactive graph analytics and visualization,'' AAAI'15: Proceedings of the Twenty-Ninth AAAI Conference on Artificial Intelligence, January 2015, Pages 4292–4293.

\bibitem{nvprofiler} Thomas Bradley, ``Gpu performance analysis and optimisation,'' NVIDIA Corporation, 2012.

\bibitem{Glorot2011} Xavier Glorot, Antoine Bordes and Yoshua Bengio, ``Deep Sparse Rectifier Neural Networks,'' Proceedings of the Fourteenth International Conference on Artificial Intelligence and Statistics, PMLR 15:315-323, 2011.

\bibitem{Narayan1997} Sridhar Narayan, ``The generalized sigmoid activation function: competitive supervised learning,'' in Information Sciences: an International Journal. Volume 99. Issue 1-2. June 1997. pp 69–82. doi: https://doi.org/10.1016/S0020-0255(96)00200-9

\bibitem{Gilmer2020} Justin Gilmer, Samuel S. Schoenholz, Patrick Riley, Oriol Vinyals and George Dahl, ``Message Passing Neural Networks,'' Machine Learning Meets Quantum Physics, pp. 199-214, 2020. Available: 10.1007/978-3-030-40245-7\_10 [Accessed 24 June 2022].

\bibitem{GraphNets} Peter W. Battaglia, Jessica B. Hamrick, Victor Bapst, Alvaro Sanchez-Gonzalez, Vinícius Flores Zambaldi, Mateusz Malinowski, Andrea Tacchetti, David Raposo, Adam Santoro, Ryan Faulkner, Çaglar Gülçehre, H. Francis Song, Andrew J. Ballard, Justin Gilmer, George E. Dahl, Ashish Vaswani, Kelsey R. Allen, Charles Nash, Victoria Langston, Chris Dyer, Nicolas Heess, Daan Wierstra, Pushmeet Kohli, Matthew M. Botvinick, Oriol Vinyals, Yujia Li, Razvan Pascanu, ``Relational inductive biases, deep learning, and graph networks,'' arxiv, 2018.

\bibitem{Spektral} Daniele Grattarola, Cesare Alippi, ``Graph Neural Networks in TensorFlow and Keras with Spektral,'' ICML 2020 - GRL+ Workshop, 2020.

\bibitem{Caffe} Yangqing Jia, Evan Shelhamer, Jeff Donahue, Sergey Karayev, Jonathan Long, Ross Girshick, Sergio Guadarrama, and Trevor Darrell. 2014. Caffe: Convolutional Architecture for Fast Feature Embedding. arXiv preprint arXiv:1408.5093 (2014).

\bibitem{Theano} Theano Development Team. 2016. Theano: A Python framework for fast computation of mathematical expressions. arXiv e-prints abs/1605.02688 (May 2016).

\bibitem{stgcn} B. Yu, H. Yin, and Z. Zhu, “Spatio-temporal graph convolutional networks:
A deep learning framework for traffic forecasting,” in Proceedings
of the 27th International Joint Conference on Artificial Intelligence
(IJCAI), 2018.

\end{thebibliography}
\end{document}